\documentclass{article} %
\usepackage{iclr2027_conference,times}

\usepackage{amsmath,amsfonts,bm}

\def\eqref#1{equation~\ref{#1}}

\def\1{\bm{1}}

\def\ry{{\textnormal{y}}}

\def\vb{{\bm{b}}}

\def\vh{{\bm{h}}}

\def\vx{{\bm{x}}}
\def\vy{{\bm{y}}}
\def\vz{{\bm{z}}}

\def\mH{{\bm{H}}}

\def\mW{{\bm{W}}}

\DeclareMathAlphabet{\mathsfit}{\encodingdefault}{\sfdefault}{m}{sl}
\SetMathAlphabet{\mathsfit}{bold}{\encodingdefault}{\sfdefault}{bx}{n}

\def\sG{{\mathbb{G}}}

\def\sW{{\mathbb{W}}}

\newcommand{\E}{\mathbb{E}}

\newcommand{\R}{\mathbb{R}}

\newcommand{\Cov}{\mathrm{Cov}}

\usepackage{hyperref}
\usepackage{url}
\usepackage{graphicx}
\usepackage{booktabs}
\usepackage{longtable}
\usepackage{amsmath,amssymb}
\usepackage{xcolor}

\usepackage{placeins}
\usepackage[normalem]{ulem}
\usepackage{makecell}

\title{When, Not How Much: Evaluating Time-Series Foundation Models on Sparse Events}

\author{%
  Daniel Schoess\\
  ETH Zurich\\
    \AND 
  Florian von Wangenheim \\
  ETH Zurich\\
}

\iclrfinalcopy %
\begin{document}

\maketitle
\fancyhead[L]{Preprint}

\begin{abstract}
Pretrained time-series foundation models (TSFMs) are evaluated as forecasters of future values, yet for sparse series many decisions depend only on which future periods contain activity.
Standard benchmarks do not assess this.
On five sparse datasets, we rank positions within forecast windows that contain both events and zeros.
The released point forecasts of 12 TSFMs improve chance-corrected average precision over training-free references by at most 0.031, and in chance-corrected AUC the median TSFM falls below them on every dataset.
With event supervision, linear probes of six frozen backbones improve on their backbone's point forecast in 29 of 30 backbone--dataset pairs.
Averaging the predicted quantiles instead of taking their median improves the ranking of most TSFMs that forecast the median, and on two datasets the strongest such outputs rival the probes.
The probes' advantage over raw-context learners depends on the dataset, and under the same probe, pretrained features outperform randomly initialized ones for five of six backbones.
For sparse-event ranking, released point forecasts thus add little over simple references, whereas lightweight event heads on frozen TSFMs rank events better than these forecasts, and the best of them exceed gradient-boosted trees trained on the raw context on three of the five datasets.
More broadly, assessing pretrained forecasters on tasks beyond value forecasting requires reporting their outputs, supervised probes of their representations, and raw-context and randomized controls side by side, since each supports a different conclusion.
\looseness=-1
\end{abstract}

\section{Introduction}
\label{sec:intro}

\begin{figure}[tb]
    \centering
    \includegraphics[width=0.9\textwidth]{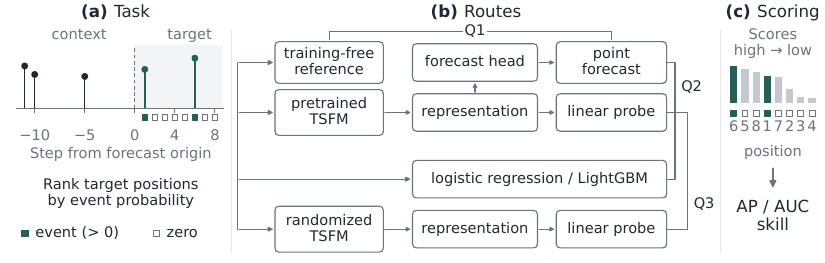}
    \caption{%
    (a) Given the context of a sparse series, rank the positions of the target window by their chance of an event (8 positions shown; the study
    uses 64).
    (b) All routes read the same context. Q1 compares the point forecasts of 12 frozen TSFMs with a training-free reference; Q2 compares linear
    probes of six frozen backbones with their point forecasts and with raw-context learners; Q3 compares each probe with the same probe on a randomized copy of its backbone.
    Probes and raw-context learners are trained on event labels.
    (c) Each route's scores are ranked within the window and scored by chance-corrected AP and AUC skill, averaged over windows.}
    \label{fig:schematic}
\end{figure}

Sparse time series arise wherever activity is intermittent.
Item-level retail sales, maintenance requests, case counts of rare conditions, pickups at peripheral transit locations, and views of niche web pages consist mostly of zeros, with occasional positive values.
In such series, the occurrence of activity carries information of its own, separate from its size.
Many decisions depend on occurrence alone: an inspection, a staffed shift, or a service visit incurs the same cost whatever the size of the event it serves.
When such resources are limited, the relevant question becomes which future periods are most likely to see activity.
We formalize this question as \emph{within-window event ranking}: given the history of a series, rank the positions of the next window by their probability of containing an event (Figure~\ref{fig:schematic}).

Pretrained time-series foundation models (TSFMs) forecast unseen series without task-specific training, and large benchmarks evaluate them as zero-shot forecasters of future values \citep{aksu2024gifteval,cohen2025boom}.
These benchmarks do not assess sparse-event ranking: zero-heavy windows form a minority of their targets (Appendix~\ref{app:benchmark-coverage}), and their metrics score forecast values rather than the ordering of events within a window.

Value forecasts also provide no direct score for occurrence.
For a nonnegative target $\ry$ with context $\vx$, the conditional mean factorizes as
\[
\E[\ry\mid \vx]=P(\ry>0\mid \vx)\,\E[\ry\mid \ry>0,\vx].
\]
Ordering positions by the mean thus ranks the expected amount of activity, which combines the probability of an event with its expected size.
Decisions triggered by any event depend on the first factor alone.
Under a Poisson or a negative binomial model with fixed dispersion, the event probability increases with the mean, so the two rankings coincide.
In general, they diverge only where a position with a higher event probability has a smaller expected event size.
Many TSFMs, however, report the median of their forecast distribution as the point forecast, the forecast that absolute-error metrics reward \citep{kolassa2016evaluating}.
For a nonnegative target, that median is zero whenever the event probability is below 50\%, so a median point forecast cannot distinguish a position with a 1\% event probability from one with 49\%.
How a forecast distribution is reduced to a point therefore affects the ranking.
Output-level evaluation also leaves open how well the frozen models rank events once event labels are available, and whether this depends on pretraining.

We address these questions on five sparse datasets from retail, e-commerce, transport, and web traffic.
First, we score the point forecasts of 12 TSFMs against a training-free reference: the seasonal profile or horizon ramp that scores best on validation data.
We measure ranking quality with average precision and AUC, each corrected for chance within the window (\emph{AP skill} and \emph{AUC skill}): zero matches a random ordering, and one is a perfect ranking.
Second, for six backbones, we train linear probes on frozen representations with event labels, a lightweight event head that measures how well the frozen features rank events under supervision.
Three controls interpret these probes.
Logistic regression on the raw context shares the learner family of the probe and so compares the frozen features with the raw context under the same learner.
LightGBM on the raw context provides a strong direct learner as a reference level.
Randomly initialized copies of each backbone test whether the gains require pretrained weights, since trained probes can extract strong features from untrained networks \citep{dempster2020rocket,zhang2018language}.

The point forecasts of the TSFMs exceed the training-free reference by at most 0.031 AP skill, and under AUC skill the median TSFM falls below it on every dataset.
Linear probes of six of these frozen models improve on the point forecasts in 29 of 30 backbone--dataset pairs and exceed raw-context logistic regression in 22 of 30, whereas their comparison with raw-context LightGBM depends on the dataset.
Under the same probe, pretrained features outperform randomized ones for five of six backbones; for Timer, randomized features score higher on three of five datasets.

Our contributions are as follows.
\begin{itemize}
\item \textbf{Task and protocol.} We define within-window event ranking for sparse series, curate five sparse evaluation datasets from public data, and score rankings with chance-corrected, tie-aware AP and AUC skill.
\item \textbf{Output-level assessment.} Point forecasts of 12 TSFMs give at most small gains over training-free references on all five datasets, and a sensitivity analysis shows that the ranking depends on how the forecast distribution is reduced to a point.
\item \textbf{Representation-level assessment with supervised and randomized controls.} Linear probes of six frozen backbones improve on their backbone's point forecast in nearly every case. Raw-context learners show that the advantage over direct event learning depends on the dataset, and randomly initialized backbones show that the advantage of pretrained weights depends on the backbone.
\end{itemize}
We conclude that evaluations of TSFMs on sparse events should report outputs, frozen-representation probes, and raw-context and randomized controls side by side.

\section{Related work}
\label{sec:related}

\paragraph{Occurrence in sparse time series.}
Several count-data models give zeros a separate component.
Hurdle models separate whether a count is zero or positive from the distribution of the positive counts \citep{mullahy1986hurdle}, and zero-inflated models mix a count distribution with a point mass at zero \citep{lambert1992zip}.
A well-studied forecasting setting with sparse series is intermittent demand, where many periods record no sales.
Croston's method smooths demand sizes and intervals separately \citep{croston1972}, and the TSB method replaces the interval with a demand probability \citep{teunter2011intermittent}.
Deep renewal processes model interarrival times and positive sizes \citep{turkmen2021}, and SPADE-S routes very sparse series to a separate arm of a neural forecaster \citep{wolff2025spades}.
These methods forecast an amount per period, even when they model occurrence internally.
Absolute-error measures such as the MAE are minimized by the median of the predictive distribution \citep{kolassa2016evaluating}, which is zero whenever zero has probability above one half.
In concurrent work on intermittent demand, \citet{chin2026accuracy} find that more accurate forecasters tend to achieve lower order-level service.
We instead score how well a method ranks event occurrences within a window.
Temporal point processes predict the times of labeled events, and HoTPP scores long-horizon event forecasts with a temporally tolerant AP \citep{karpukhin2026hottp}.
Our targets lie on a fixed grid, so every position carries a label and no matching tolerance is needed.
\looseness=-1

\paragraph{Evaluating pretrained forecasters.}
GIFT-Eval and BOOM benchmark TSFMs as zero-shot forecasters with point and probabilistic error metrics \citep{aksu2024gifteval,cohen2025boom}.
Zero-heavy windows form a minority of the forecast windows in both (Section~\ref{sec:populations}), and their per-step metrics do not score how events are ordered within a window.
A second line of work asks what frozen TSFM representations encode rather than what the models forecast.
Prior work trains classifiers on frozen TSFM embeddings \citep{auer2025pretrained,feofanov2026mantis} and locates synthetic concepts or generator states with layer-wise linear probes \citep{wilinski2025exploring,chemeris2026aionoscope}.
These studies probe for labels other than the forecast target.
We probe for the occurrence of the forecast target and score the probe and the released forecast of the same frozen model with the same ranking metric, so the two can be compared directly.
\looseness=-1

\paragraph{Controls for probes.}
A probe score alone does not show what the representation contributes.
Control tasks measure how much of a probe's accuracy comes from the probe itself rather than the representation \citep{hewitt2019designing}.
\citet{belinkov2022probing} also discusses random baselines, which train the same probe on a randomized version of the model, and notes that even random features carry information the probe can decode.
A linear classifier on features pooled from random convolutional kernels matched the most accurate time-series classifiers of its time \citep{dempster2020rocket}.
Frozen, randomly initialized LSTMs came within a few percentage points of trained language-model encoders on syntactic tagging when the probe used all labeled data, but their accuracy dropped far more than that of trained encoders when the probe's training data were reduced \citep{zhang2018language}.
In one LLM-based forecaster, replacing the pretrained language model with a frozen, randomly initialized one, while still training the layers around it, leaves forecast error close to that of the pretrained variants \citep{tan2024language}.
The TSFM probing and frozen-feature studies above do not report a randomized-backbone control.
We therefore report each probe next to the same learner on the raw context and the same probe on a randomized backbone, together with the absolute scores of both.

\section{Three questions on forecasts, representations, and pretraining}
\label{sec:setup}

We evaluate each pretrained model through three questions, each answered by paired comparisons on the same windows (Table~\ref{tab:design}).
First, how well does the point forecast of a TSFM rank events within a window, measured against a training-free summary of the same context?
Second, does an event-supervised probe of the frozen representation rank better than that output, and better than the same learner trained on the raw context?
Third, does the probe rank events better with pretrained weights than with a randomly initialized copy of the same backbone?
Appendices~\ref{app:data} and~\ref{app:provenance} give the full protocol.

\begin{table}[tb]
\caption{The three evaluation questions and the paired comparison that answers each, on all five datasets.
Event labels: which of the two compared scores is trained on the event labels of the training split.
The Q1 reference is selected from four or six candidates on validation event labels.
The linear probe is the primary probe; the GBM probe, LightGBM on the frozen representation, is a sensitivity analysis.}
\label{tab:design}
\begin{center}
\footnotesize
\begin{tabular}{@{}llll@{}}
\toprule
Question & Compared scores & Event labels & Models \\
\midrule
Q1 Output & point forecast vs.\ training-free reference & neither & 12 TSFMs \\
\addlinespace[2pt]
Q2 Probe & linear probe vs.\ point forecast & probe & 6 backbones \\
 & linear probe vs.\ raw-context logistic regression & both & 6 backbones \\
 & linear probe vs.\ raw-context LightGBM & both & 6 backbones \\
 \addlinespace[2pt]
Q3 Pretraining & pretrained vs.\ randomized linear probe & both & 6 backbones \\
\bottomrule
\end{tabular}
\end{center}
\end{table}

\subsection{Task, datasets, and windows}
\label{sec:populations}

Zero-heavy targets form a minority of standard forecasting benchmarks: 24\% of GIFT-Eval \citep{aksu2024gifteval} and 1.6\% of BOOM \citep{cohen2025boom} forecast windows contain at least 40\% zeros (Appendix~\ref{app:benchmark-coverage}).
We therefore build five datasets of nonnegative sparse series from raw public sources: daily customer purchases (Amazon) \citep{berke2024open_amazon_data}, daily store--item sales (Favorita) \citep{favorita-grocery-sales-forecasting} and M5 \citep{MAKRIDAKIS20221325_m5}, hourly pickups per location (NYC taxi) \citep{nyctlc2024}, and daily Wikipedia page views (web traffic) \citep{godahewa_2022_7371038}.
An event is a position with a positive observed value.
We keep series with at least two events and at least 40\% zero positions (Appendix~\ref{app:data}).
\looseness=-1

A window $w$ pairs a context $\vx_w\in\R^{C}$ of $C=364$ daily or $C=336$ hourly observations, one year of days or two weeks of hours, with the $H=64$ target positions that follow it.
We set $H$ to the shortest native output limit among the evaluated TSFMs.
The target has labels $\vy_w\in\{0,1\}^{H}$, where $y_{w,h}=1$ if position $h$ contains an event.
Windows are tiled backwards from the end of the data with stride $H$, so targets do not overlap within a series.
A window enters the study if its context contains at least two events.
It is scored only if its target contains $k_w=\sum_h y_{w,h}$ events with $0<k_w<H$, because AP is undefined without positives and the skill normalization of Section~\ref{sec:metrics} is undefined without negatives.
All results therefore concern targets with mixed labels; they do not measure whether a method recognizes an entirely inactive horizon.
For the four daily datasets, five seeded splits partition the series into training, validation, and test sets, so trained models are scored on unseen series.
NYC taxi has too few locations for series-disjoint splits (164 evaluated) and is split by year instead.
Table~\ref{tab:population} summarizes the evaluated windows and series.

\begin{table}[tb]
\caption{Evaluation datasets. Windows: evaluated test windows, pooled over splits. 
Series: unique series contributing test windows. 
Zeros: share of zero positions in the evaluated targets. All windows use a 64-step horizon and stride.}
\small
\label{tab:population}
\begin{center}
\begin{tabular}{llllrrr}
\toprule
Dataset & Cadence & Context & Split design & Windows & Series & Zeros (\%) \\
\midrule
Amazon & daily & 364 & 5 series-disjoint & 22,600 & 1,135 & 87.5 \\
Favorita & daily & 364 & 5 series-disjoint & 321,747 & 26,427 & 45.7 \\
M5 & daily & 364 & 5 series-disjoint & 120,063 & 5,979 & 64.1 \\
NYC taxi & hourly & 336 & 1 calendar & 19,207 & 164 & 73.7 \\
Web traffic & daily & 364 & 5 series-disjoint & 41,816 & 1,597 & 65.6 \\
\bottomrule
\end{tabular}

\end{center}
\end{table}

\subsection{Point forecasts, frozen representations, and raw contexts}
\label{sec:references}

\paragraph{Point forecasts.}

We evaluate the largest publicly available checkpoint of twelve TSFMs: Chronos-2 \citep{ansari2025chronos2}, Chronos-Bolt \citep{ansari2024chronos}, Moirai-1.1 \citep{woo2024moirai}, Moirai-2.0 \citep{liu2025moirai2}, Moirai-MoE \citep{liu2024moiraimoe}, Sundial \citep{liu2025sundial}, TimeMoE \citep{shi2025timemoe}, Timer \citep{liu2025timerxl}, TimesFM-2.5 \citep{timesfm}, TimesFM-3 \citep{timesfm}, TiRex \citep{auer2025tirex}, and Toto-2.0 \citep{datadog2026toto2}.
Each TSFM ranks positions by its point forecast, the output these models are released and used with: a median for seven TSFMs, and a mean of samples or a model point output for the other five (Table~\ref{tab:point-forecast}).
An approximate event probability derived from the quantiles serves as a sensitivity analysis for the ten TSFMs that output quantiles or samples (Appendix~\ref{sec:quantiles}).
To test whether a forecast ranks positions better than a simple summary of the same context, we also score a training-free reference: the seasonal event-rate or mean profile, or a horizon ramp, selected on each split by validation AP skill (Appendix~\ref{app:impl-references}).
We do not use Croston's method or TSB \citep{croston1972,teunter2011intermittent} as references, because their forecast is constant across the horizon, so their within-window AP and AUC skill is zero.
\looseness=-1

\paragraph{Frozen representations.}
All TSFMs stay frozen throughout, which keeps the weights identical across the point forecast, the probe, and the pretrained side of the randomized comparison.
Ten TSFMs produce the full 64-step horizon in one forward pass (Appendix~\ref{app:impl-tsfm}), and we probe six of them: Chronos-2, Chronos-Bolt, Sundial, TimeMoE, Timer, and Toto-2.0.
They span encoder-only, encoder--decoder, and decoder-only architectures, including a mixture-of-experts backbone and a flow-matching head.
We fixed this subset before training any probe.
It omits Moirai-1.1, Moirai-2.0, TimesFM-2.5, and TimesFM-3, which include the strongest point forecasters.
Each probe reads a forecast-aligned representation (Appendix~\ref{app:impl-tsfm}), so it sees the same input as the forecast, and neither depends on predictions generated earlier in the horizon.

\paragraph{Raw contexts.}

Each probe has a raw-context counterpart trained with the same learner family, the same windows, and the same training sample.
Logistic regression maps the context to the 64 position scores and serves as the same-family control for the linear probe.
LightGBM \citep{ke2017lightgbm} fits one classifier per position and shows what nonlinear event learning attains directly from the same history (Appendix~\ref{app:impl-training}).

\paragraph{Randomized backbones.}

A trained probe can extract strong features from an untrained network (Section~\ref{sec:related}), so a probe gain over the forecast does not by itself show that pretraining contributes.
We therefore replace the pretrained weights of each backbone with the default initialization of its source code, keeping architecture, preprocessing, and extraction path unchanged (Appendix~\ref{app:impl-tsfm}); each split draws its own initialization.
The linear probe is then trained with the same procedure and seed as for the pretrained backbone.
Because initialization schemes differ across sources, we compare pretrained and randomized features within a backbone, not randomized features across backbones.
We also repeat the GBM probe on randomized copies of Chronos-2 and Toto-2.0 on the daily datasets as a secondary check with a nonlinear probe (Appendix~\ref{app:readout-sensitivity}).
Since the pretrained-minus-randomized gap can depend on the amount of probe training data \citep{zhang2018language}, which we cap at 100{,}000 windows per split, we verify for Chronos-2 and Timer that larger training sets change AP skill by at most 0.0054 and leave the sign of every such comparison unchanged (Appendix~\ref{app:training-size}).
\looseness=-1

\subsection{Scoring and uncertainty}
\label{sec:metrics}
Each method assigns scores $\vz_w\in\R^{H}$ to the target positions of window $w$, and we rank and score each window on its own, as information retrieval scores each query \citep{manning2008ir}.
Raw AP depends on prevalence and ties, so we report
\[
S^{\mathrm{AP}}_w=\frac{\mathrm{AP}_w-\mathrm{AP}^0_w}{1-\mathrm{AP}^0_w},
\]
where $\mathrm{AP}^0_w$ is the expected AP of window $w$ when its labels are permuted uniformly at random with its scores and positive count held fixed (Appendix~\ref{app:metrics}).
Tied positions enter AP together at their shared threshold rather than being averaged over tie orderings \citep{mcsherry2008ties}.
We also report $S^{\mathrm{AUC}}_w=2\,\mathrm{AUROC}_w-1$, with half credit for tied pairs.
AP emphasizes the top of the ranking, whereas AUC weights all positive--negative pairs.

Primary estimates average window-level skills with equal weight; equal-series averages are a sensitivity analysis (Appendix~\ref{app:weighting}).
We do not pool positions across windows, because a pooled ranking can reward differences in activity between windows without ordering positions within any window (Appendix~\ref{app:metrics-pooling}).
Uncertainty is reported as pointwise 95\% paired bootstrap intervals from 2{,}000 resamples of the eligible series of each dataset.
The intervals are not corrected for multiplicity, and counts of intervals that exclude zero are descriptive.
They are conditional on the fitted predictors; split and initialization seeds vary together, so they do not isolate training or initialization variability.
For the four daily datasets, we also count in how many of the five splits each paired difference keeps the sign of its pooled estimate (Table~\ref{tab:split-signs}).
NYC taxi has one split, so its intervals reflect one fitted run per procedure.
\looseness=-1

\section{Results}
\label{sec:results}

We answer the three questions of Section~\ref{sec:setup} in turn.
Figures report AP skill; the AUC-skill figures are in Appendix~\ref{app:complete}, and the text states where the two metrics differ.
Appendix~\ref{app:precision} expresses the results as the number of events among the eight highest-ranked positions of a window.
\looseness=-1

\subsection{Point forecasts give limited gains over training-free references}
\label{sec:results-references}

\begin{table}[tb]
\caption{Point forecasts of the 12 TSFMs compared with the validation-selected training-free reference, equal-window weighting. 
``Ref.'' is the reference's skill; ``Median'' and ``Max'' give the median and maximum over TSFMs of the TSFM-minus-reference difference, and ``Best probe'' the maximum over the six probed backbones of the pretrained linear probe's difference from the reference; both maxima are descriptive test-set maxima.
Per-TSFM values with paired intervals are in Figure~\ref{fig:absolute} and Tables~\ref{tab:complete-amazon}--\ref{tab:complete-web-traffic}.
}
\label{tab:output-summary}
\begin{center}
\footnotesize
\begin{tabular}{lrrrrrrrr}
\toprule
 & \multicolumn{4}{c}{AP skill} & \multicolumn{4}{c}{AUC skill} \\
\cmidrule(lr){2-5}\cmidrule(lr){6-9}
Dataset & Ref. & Median & Max & Best probe & Ref. & Median & Max & Best probe \\
\midrule
Amazon      & 0.008 & $-$0.002 & 0.002 & 0.006 & 0.024 & $-$0.017 & $-$0.006 & 0.009 \\
Favorita    & 0.103 & 0.007 & 0.023 & 0.092 & 0.114 & $-$0.012 & 0.005 & 0.080 \\
M5          & 0.043 & $-$0.005 & 0.013 & 0.033 & 0.062 & $-$0.022 & 0.001 & 0.040 \\
NYC taxi    & 0.185 & 0.013 & 0.031 & 0.049 & 0.310 & $-$0.053 & 0.020 & 0.042 \\
Web traffic & 0.015 & 0.003 & 0.006 & 0.057 & 0.016 & $-$0.003 & 0.004 & 0.071 \\
\bottomrule
\end{tabular}
\end{center}
\end{table}

Point forecasts rank events within a window little better than a training-free reference, and often worse (Table~\ref{tab:output-summary}).
No TSFM point forecast gains more than 0.031 AP skill over the reference on any dataset.
AUC skill gives a less favorable picture: the median over TSFMs falls below the reference on all five datasets.
The small AP gains also depend on the averaging unit: on web traffic, weighting each series equally instead of each window places the reference above every TSFM (Appendix~\ref{app:weighting}).

Although every scored window contains an event, outside Amazon the seven median forecasts stay below 1\% of the mean nonzero context value at all 64 positions in 11--54\% of scored windows, whereas the other five TSFMs do so in at most 2\% (Appendix~\ref{app:degeneracy}).
Both groups gain at most 0.031 AP skill over the reference, so near-zero median forecasts alone do not account for the limited gains.
Ranking by an approximate event probability derived from the quantiles does not improve on the point forecasts, whereas averaging the quantiles improves on most median forecasts and raises the largest gain over the reference from 0.031 to 0.057 AP skill (Appendix~\ref{sec:quantiles}).
Point forecasts are trained to predict amounts and use no event labels.
Section~\ref{sec:results-representation} tests whether the frozen representations behind these forecasts support better event ranking once a probe is trained on event labels.

\subsection{Linear probes outperform point forecasts}
\label{sec:results-representation}

\begin{figure}[tb]
\begin{center}
\includegraphics[width=0.9\textwidth]{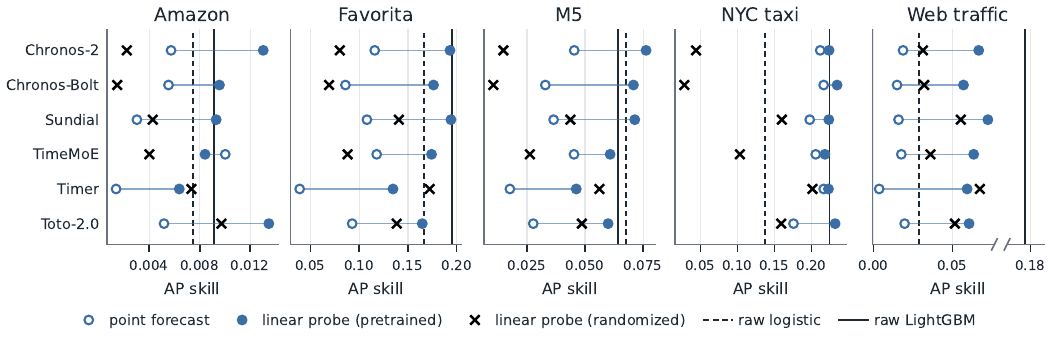}
\end{center}
\caption{AP skill of the six probed backbones on the five datasets, equal-window weighting. Axes differ by dataset; the web-traffic axis is broken. Points are marginal estimates, so their overlap is not a test; Table~\ref{tab:scoreboard} gives the paired comparisons and Tables~\ref{tab:complete-amazon}--\ref{tab:complete-web-traffic} the per-backbone intervals. Figure~\ref{fig:representation-full} adds AUC skill.}
\label{fig:representation}
\end{figure}

\begin{table}[tb]
\caption{Paired AP-skill contrasts of the six probed backbones, equal-window weighting. Each cell gives the median difference over the six backbones and, in parentheses, how many of the six paired 95\% intervals lie above zero, span zero, or lie below zero. 
Per-backbone values and AUC skill are in Tables~\ref{tab:complete-amazon}--\ref{tab:complete-web-traffic} and~\ref{tab:randomized}, and split sign counts in Table~\ref{tab:split-signs}.}
\label{tab:scoreboard}

\begin{center}

\resizebox{\linewidth}{!}{%
  \begin{tabular}{@{}lccccc@{}}
\toprule
Contrast & Amazon & Favorita & M5 & NYC taxi & Web traffic \\
\midrule
Probe $-$ point forecast & 0.006 (5/1/0) & 0.082 (6/0/0) & 0.032 (6/0/0) & 0.015 (6/0/0) & 0.047 (6/0/0) \\
Probe $-$ raw logistic & 0.002 (3/3/0) & 0.008 (4/0/2) & $-$0.002 (3/0/3) & 0.087 (6/0/0) & 0.033 (6/0/0) \\
Probe $-$ raw LightGBM & 0.000 (2/3/1) & $-$0.020 (0/0/6) & 0.002 (3/0/3) & $-$0.001 (2/3/1) & $-$0.115 (0/0/6) \\
\addlinespace
Pretrained $-$ randomized & 0.005 (5/1/0) & 0.070 (5/0/1) & 0.031 (5/0/1) & 0.094 (6/0/0) & 0.021 (5/0/1) \\
\bottomrule
\end{tabular}
}
\end{center}
\end{table}

\paragraph{Gains over the point forecast are consistent.}
A linear probe of the frozen representation ranks events better than the backbone's point forecast in 29 of 30 backbone--dataset pairs under each metric, with paired intervals above zero (Table~\ref{tab:scoreboard} and Tables~\ref{tab:complete-amazon}--\ref{tab:complete-web-traffic}).
The only exception is TimeMoE on Amazon, where both intervals span zero.
On web traffic, the gain in AP skill exceeds the forecast's own skill for all six backbones (Figure~\ref{fig:representation}).
On the four daily datasets, the difference from the point forecast has the same sign in all five seeded splits for 24 of 24 AP and 23 of 24 AUC contrasts; the exception is again TimeMoE on Amazon (Appendix~\ref{app:split-stability}).
Only the probe is trained on event labels, so we also train the same logistic learner on each backbone's outputs (Appendix~\ref{app:output-readout}).
Probes outperform these output readouts on Favorita, M5, and web traffic under both metrics.
The probes of the three median-forecasting backbones also exceed the quantile average of their own backbone in 28 of 30 comparisons (Appendix~\ref{sec:quantiles}).
On M5 and NYC taxi, the strongest quantile average among the seven median-forecasting TSFMs exceeds most of the six probes under both metrics.

\paragraph{Probes exceed raw-context logistic regression in most pairs.}
The probes also exceed logistic regression on the raw context, with intervals above zero in 22 of 30 pairs under each metric.
This holds for all six backbones on NYC taxi and web traffic, and in AP skill for three on Amazon, four on Favorita, and three on M5 (Table~\ref{tab:scoreboard}).
The other AP intervals span zero on Amazon and lie below zero on Favorita and M5.
Both learners are linear and are trained on the same windows with the same regularization grid and selection rule.
Where the probe leads, event supervision alone therefore does not account for its gain.
Unlike the probes, whose backbones rescale each window, raw logistic regression standardizes each context position with training-set statistics, which on NYC taxi come from a different year than the test set.
Fitted on test-year windows, raw logistic regression gains 0.086 AP skill there (Appendix~\ref{app:output-readout}), so the probes' median lead of 0.087 on NYC taxi does not isolate the features.
The randomized backbones keep the probes' preprocessing, and on NYC taxi pretrained features still exceed them by a median of 0.094.

\paragraph{The comparison with direct tree learning depends on the dataset.}
Raw-context LightGBM marks a level that direct event learning reaches from the same history (Table~\ref{tab:scoreboard}).
The best linear probes exceed it on Amazon, M5, and NYC taxi, by up to 0.012 AP skill, and come within 0.003 of it on Favorita, where all six probes trail it.
Web traffic is the clear exception: every linear probe trails raw LightGBM, by a median of 0.115 AP skill, and a GBM probe narrows but does not close this gap (Appendix~\ref{app:readout-sensitivity}).
On windows without missing values, the median gap shrinks to 0.039 (Appendix~\ref{app:web-missing}).
None of the tested methods achieves substantial event-ranking skill on Amazon: all stay below 0.014 AP skill and place fewer than 0.1 more events than a random ordering among the eight highest-ranked positions (Appendix~\ref{app:precision}).
Under equal-series weighting, no probe interval against a point forecast or raw learner moves from one side of zero to the other, and outside Amazon six linear-probe intervals against raw LightGBM then span zero (Table~\ref{tab:readout-weighting}).

\subsection{The advantage of pretrained features depends on the backbone}
\label{sec:results-randomized}

\begin{figure}[tb]
\begin{center}
\includegraphics[width=0.9\textwidth]{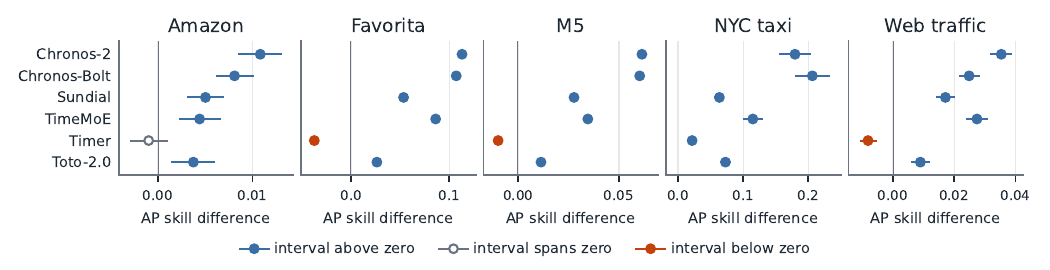}
\end{center}
\caption{Pretrained-minus-randomized difference in AP skill under the linear probe, with pointwise 95\% paired bootstrap intervals over series, often narrower than the markers; equal-window weighting. NYC taxi has one fitted run per initialization. Figure~\ref{fig:randomized-full} adds AUC skill.}
\label{fig:randomized}
\end{figure}

\paragraph{Pretrained features lead for five of six backbones.}
Under the linear probe, pretrained features outperform randomized ones in 26 of 30 AP comparisons and 25 of 30 AUC comparisons (Figure~\ref{fig:randomized}, Table~\ref{tab:scoreboard}).
Timer accounts for all four AP exceptions and four of the five AUC exceptions.
Chronos-2, Chronos-Bolt, Sundial, and TimeMoE are ahead in all ten comparisons, and Toto-2.0 in nine, with its Amazon AUC interval spanning zero.
The size of this advantage varies widely: on Favorita, the AP-skill difference is 0.113 for Chronos-2 and 0.107 for Chronos-Bolt, but 0.053 for Sundial and 0.026 for Toto-2.0.

\paragraph{Randomized features already attain substantial skill.}
On every dataset and metric, randomized-probe skill varies more across the six backbones than pretrained-probe skill.
In AP skill, pretrained probes span 0.135--0.194 on Favorita and 0.218--0.235 on NYC taxi, whereas randomized probes span 0.069--0.172 and 0.028--0.201.
Randomized probes exceed their backbone's point forecast in 15 of 30 pairs under AP and 16 under AUC skill (point estimates), including all six on web traffic and Sundial, Timer, and Toto-2.0 on every daily dataset.
\looseness=-1

\paragraph{Timer is the exception.}
For Timer, randomized features outperform pretrained ones under the linear probe on Favorita, M5, and web traffic under both metrics, with the same sign in all five seeded splits (Table~\ref{tab:split-signs}); under equal-series weighting, the web-traffic AP difference becomes positive (Appendix~\ref{app:weighting}).
On Favorita, the randomized probe reaches 0.172 AP skill and the pretrained probe 0.135.
Only on NYC taxi are pretrained Timer features ahead (0.022 AP skill).
Timer's point forecast is also the weakest of the six on every daily dataset under both metrics, but among the strongest on NYC taxi.
Appendix~\ref{app:timer} reports the remaining comparisons.

\paragraph{The lead over LightGBM needs pretrained weights.}
Outside Amazon, no randomized probe reaches raw LightGBM under either metric.
On M5 and NYC taxi, the probes' advantage over raw LightGBM therefore appears only with pretrained weights.
On M5, no randomized probe reaches even raw logistic regression.
Randomized probes do exceed raw logistic regression in several other cases, including all six on web traffic and Timer on Favorita (0.172 versus 0.167).
\looseness=-1

\section{Discussion and limitations}
\label{sec:discussion}

\paragraph{Interpreting the probe gains.}
The raw-logistic and randomized controls train both sides on the same event labels.
Hence, they test the representation under matched supervision: the probe leads raw-context logistic regression in 22 and the randomized probe in 26 of 30 AP comparisons.
On M5 and NYC taxi, the best linear probes also exceed raw-context LightGBM, and no randomized probe does.
On M5, the three probes that exceed raw LightGBM do so by 0.007--0.012 AP skill, or 10--19\% of its skill, with the same sign in all five seeded splits.

\paragraph{Why point forecasts rank events poorly.}
A point forecast targets an amount: the mean mixes event probability with event size, and the median of a zero-heavy distribution is zero wherever an event is less likely than not (Section~\ref{sec:intro}).
Near-zero median forecasts (Section~\ref{sec:results-references}) are consistent with this second effect, but the other five TSFMs rarely forecast near zero and gain equally little.
Averaging the quantiles instead of taking their median improves most median forecasts, and on M5 and NYC taxi the strongest quantile average exceeds most probes, so the ranking also depends on how the forecast distribution is reduced to a point (Appendix~\ref{sec:quantiles}).
Output heads that forecast both factors of the decomposition in Section~\ref{sec:intro}, the event probability and the positive amount, are a direct next step.

\paragraph{Where direct learning and random features match or exceed the probes.}
Web traffic is the one dataset on which all probes trail raw LightGBM by a wide margin, even with a GBM readout.
Randomized probes also reach substantial skill (Section~\ref{sec:results-randomized}), in line with strong random-feature baselines in time-series classification \citep{dempster2020rocket}.
Reporting only the pretrained probe would therefore credit pretraining with skill that randomized features already reach. 
Timer is the extreme case: its randomized features outperform its pretrained ones on Favorita, M5, and web traffic.
To look for a cause, we fit linear readouts of the recent context from pretrained and randomized features (Appendix~\ref{app:timer}).
On all five datasets, Timer's pretrained features encode the recent event rate better than its randomized ones, but the recent event pattern worse. No other backbone shows this pattern loss on every dataset; Toto-2.0 shows a smaller one on three.
Across backbones, this pattern loss tracks the pretrained-minus-randomized probe gap, but it does not explain which of Timer's datasets reverse, so it is a correlate rather than an explanation.

\paragraph{Recommended reporting.}
We recommend that evaluations of TSFMs on sparse events report four comparisons side by side, each for the conclusion it supports.
A training-free reference measures what the point forecast adds to event ranking.
The probe's own learner trained on the raw context tests the frozen features under matched supervision, and a strong direct learner tests whether the probe is competitive with direct event learning.
A randomized backbone under the same probe tests the pretrained weights against a fixed initialization.
Each paired difference should appear next to the absolute scores of both routes.
At the output level, the ranking metric and the averaging unit also change conclusions (Section~\ref{sec:results-references}), so both should be stated.
In current benchmarks, zero-heavy windows form a minority and metrics score forecast values (Section~\ref{sec:populations}); our five datasets and scoring protocol offer a starting point for an occurrence-ranking track.

\paragraph{Limitations.}

Our conclusions cover five public datasets, target-only inputs, a 64-step horizon, and windows whose target contains both events and zeros, so we do not assess whether a method recognizes a fully inactive horizon.
Events are recorded positive values, so zeros can also reflect missing records, as for 68\% of the web-traffic zero labels, or unavailability, as for Favorita items that may have been out of stock (Appendix~\ref{app:data-zeros}).
Daily splits separate series but share calendar time, which can favor the trained routes over the point forecasts but applies equally to probes and raw-context learners.
Some TSFMs disclose pretraining sources that overlap with our data, for web traffic including our pages and dates (Appendix~\ref{app:data-overlap}), which could favor pretrained over randomized features.
In the 12 backbone--dataset pairs with disclosed exclusion, mostly on Amazon and M5, pretrained features exceed randomized copies by a median of 0.019 AP skill, with 11 of 12 intervals above zero, against 0.027 over all 30 pairs (Table~\ref{tab:disclosed-exclusion}).
The probes that exceed raw LightGBM on M5 also all come from backbones that disclose excluding M5.
Probes cover six of the ten TSFMs that meet our extraction criterion, and each representation is read at one extraction point.
Probe results depend on extraction and aggregation choices \citep[App.~C.2]{auer2025pretrained}, so other extraction points or other backbones may rank events better.
Event-supervised fine-tuning, outside our frozen-backbone scope, could raise the skill of both TSFM routes.
Raw-context LightGBM also does not represent all direct learners; feature-engineered or dedicated event and count models may rank events better.
Our intervals are pointwise and uncorrected for multiplicity, and NYC taxi contributes one fitted run per procedure.
Our metrics assess ranking only, not the calibration of event probabilities or their value for downstream decisions.

\bibliography{references}
\bibliographystyle{iclr2027_conference}

\appendix

\FloatBarrier
\section{Datasets and evaluation windows}
\label{app:data}

\subsection{Sources and series construction}
\label{app:data-sources}

Table~\ref{tab:data-sources} lists the source, series unit, and period of each dataset.

\begin{table}[tb]
  \caption{Sources and series construction. The Amazon purchase amount is quantity times unit price, summed per day. Retained: series that pass the sparsity filter. For NYC taxi, counts are given for 2022/2023/2024, since the filter is applied within each year.}
  \label{tab:data-sources}
  \begin{center}
  \resizebox{\textwidth}{!}{\begin{tabular}{lllllrr}
\toprule
Dataset & Source & Series & Value per position & Period & Source series & Retained \\
\midrule
Amazon & \citep{berke2024open_amazon_data} & respondent & daily purchase amount & 2018-01-01 to 2024-08-15 & 5{,}027 & 5{,}024 \\
Favorita & \citep{favorita-grocery-sales-forecasting} & store--item pair & daily unit sales & 2013-01-01 to 2017-08-15 & 174{,}685 & 121{,}619 \\
M5 & \citep{MAKRIDAKIS20221325_m5} & item--store pair & daily unit sales & 2011-01-29 to 2016-05-22 & 30{,}490 & 26{,}334 \\
NYC taxi & \citep{nyctlc2024} & pickup location & hourly pickups & 2022-01-01 to 2024-12-31 & 262/263/263 & 185/188/168 \\
Web traffic & \citep{godahewa2021monash,godahewa_2022_7371038} & page & daily page views & 2015-07-01 to 2022-06-30 & 145{,}063 & 7{,}048 \\
\bottomrule
\end{tabular}
}
  \end{center}
  \end{table}

We retain series with at least two active positions and at least 40\% inactive positions.
The inactive share is computed over the dataset's common calendar range for Amazon, Favorita, and M5, over each calendar year for NYC taxi, and over the series length for web traffic.

\subsection{What a zero means}
\label{app:data-zeros}
The pipeline stores each series as its nonzero records and places them on a zero-initialized regular grid, so every position without a record is zero.
A zero therefore means that the source records no activity at that position; whether this reflects true inactivity depends on the source.
Positions before the first recorded activity of a series, for example before an item is introduced, are also zero.
The Favorita source omits store--item--days with zero unit sales and records no information on whether an item was in stock.
Unit sales can be fractional for items sold by weight; the pipeline truncates them to integers before aggregation, so sales below one unit (0.4\% of source records) count as zeros, as do returns.
The header of the web-traffic source file states that missing values in the original data ``have been simply replaced by zeros''.
The version of the same dataset with missing values \citep{godahewa_2022_7370977} marks them explicitly.
Missing values make up 55\% of all web-traffic positions and 79\% of its zeros.
In the scored targets, 45\% of positions and 68\% of the zero labels are missing, and 91\% of windows have at least one missing target day.
Web-traffic results therefore concern the ranking of recorded activity.
Because missing views enter every method as zeros, no method can distinguish them from inactivity.
Appendix~\ref{app:web-missing} repeats the main comparisons on windows without missing values.
After preprocessing, every evaluated target position has a value; in web traffic, these values include missing page views stored as zeros.

\subsection{Splits, windows, and eligibility}
\label{app:data-windows}

For the daily datasets, seeds 42--46 each partition series into training, validation, and test sets (90/5/5\%), stratified by the number of active positions in 20 quantile bins.
A series can enter the test sets of several seeds, so Table~\ref{tab:population} counts unique series.
NYC taxi uses training year 2022, validation year 2023, and test year 2024.
All 168 locations retained in 2024 are also retained in 2022 and 2023, which contain 17 and 20 additional locations.
A window consists of 364 daily or 336 hourly context positions followed by 64 target positions.
Windows are tiled backwards from the end of each split's grid with stride 64, so targets do not overlap within a series, whereas contexts may.
For NYC taxi, the last target ends 48 hours before the end of 2024.
Before inference, a window is retained if its context contains at least two positive observations.
For evaluation, its target must additionally contain $k_w$ positives with $0<k_w<64$: AP is undefined without positives, and AP skill is undefined when all targets are positive.
All scores and paired contrasts within a dataset use this one set of windows.

\begin{table}[tb]
\caption{Retention of test windows, pooled over splits, from candidate windows through the context and mixed-label filters.
Target-zero percentages pool valid target positions across represented splits.}
\label{tab:population-retention}
\begin{center}
\resizebox{\textwidth}{!}{\begin{tabular}{llrrrrrrrrr}
\toprule
 & & & & \multicolumn{3}{c}{Windows} & \multicolumn{2}{c}{Series} & \multicolumn{2}{c}{Target zeros (\%)} \\
\cmidrule(lr){5-7} \cmidrule(lr){8-9} \cmidrule(lr){10-11}
Dataset & Cadence & Context & Splits & Candidate & Context-eligible & Mixed-label & Candidate & Mixed-label & Candidate & Mixed-label \\
\midrule
Amazon & daily & 364 & 5 & 30,240 & 26,914 & 22,600 & 1,154 & 1,135 & 90.6 & 87.5 \\
Favorita & daily & 364 & 5 & 608,100 & 375,861 & 321,747 & 27,500 & 26,427 & 67.9 & 45.7 \\
M5 & daily & 364 & 5 & 158,040 & 130,032 & 120,063 & 5,979 & 5,979 & 72.0 & 64.1 \\
NYC taxi & hourly & 336 & 1 & 22,008 & 20,440 & 19,207 & 168 & 164 & 77.0 & 73.7 \\
Web traffic & daily & 364 & 5 & 60,010 & 50,612 & 41,816 & 1,615 & 1,597 & 70.1 & 65.6 \\
\bottomrule
\end{tabular}
}
\end{center}
\end{table}

Candidate windows lie on each study grid after dataset-level preprocessing.
Context-eligible windows contain at least two positive context observations; mixed-label windows additionally contain both target classes and form the evaluation dataset.
Table~\ref{tab:population-retention} reports all three stages, the corresponding series counts, and candidate and evaluated target-zero rates.

\subsection{Zero-heavy targets in forecasting benchmarks}
\label{app:benchmark-coverage}
We audited the official forecast targets of all 55 GIFT-Eval source configurations (dataset--frequency pairs; 97 configurations across horizons) and all 2{,}807 BOOM series at their pinned repository and data revisions.
Forecast windows follow the official rolling evaluation protocol shared by both benchmarks.
Table~\ref{tab:external-coverage} reports the share of all forecast windows, pooled across configurations, whose finite targets contain at least the indicated fraction of exact zeros.
The other columns first compute the share of valid target points or absolute target mass in qualifying windows within each configuration and then weight configurations equally.
Nonfinite values are excluded from zero-fraction denominators.
The BOOM shares barely change across thresholds, so nearly all of its zero-heavy windows contain at least 90\% zeros.
Windows with at least 80\% zeros hold 3.96\% of GIFT-Eval target points but only 0.28\% of its absolute target mass, which normalizes the benchmark's CRPS.
For comparison, 73 to over 99\% of the candidate windows of our five datasets (Appendix~\ref{app:data-windows}) meet the 40\% threshold and 46--85\% the 80\% threshold.
Candidate windows are the comparable set, because the benchmark shares also include all-zero targets.

\begin{table}[h]
\caption{Coverage of exact-zero-heavy forecast windows in the audited official benchmark targets. Values are percentages.}
\label{tab:external-coverage}
\begin{center}
\small
\begin{tabular}{llrrr}
\toprule
Benchmark & Zero threshold & Windows & Valid points & Absolute target mass \\
\midrule
GIFT-Eval & 40\% & 24.19 & 12.30 & 6.74 \\
GIFT-Eval & 80\% & 12.02 & 3.96 & 0.28 \\
GIFT-Eval & 90\% & 8.69 & 3.27 & 0.11 \\
BOOM & 40\% & 1.55 & 0.87 & 0.30 \\
BOOM & 80\% & 1.55 & 0.86 & 0.30 \\
BOOM & 90\% & 1.55 & 0.86 & 0.30 \\
\bottomrule
\end{tabular}
\end{center}
\end{table}

\subsection{Pretraining-source overlap}
\label{app:data-overlap}
Table~\ref{tab:checkpoint-overlap} classifies, for each TSFM and dataset, what the TSFM's own documentation discloses about the dataset's source.
These classifications draw on the model cards and technical reports of Chronos \citep{ansari2024chronos,ansari2025chronos2}, Toto \citep{datadog2026toto2}, TimesFM \citep{timesfm}, Moirai \citep{woo2024moirai,liu2024moiraimoe,liu2025moirai2}, TiRex \citep{auer2025tirex}, TimeMoE \citep{shi2025timemoe}, Sundial \citep{liu2025sundial}, and Timer \citep{liu2025timerxl}.
Moirai-1.1 is U throughout because its release documentation does not describe its training data.
Chronos-Bolt is U except for M5, which its model card lists among benchmark datasets not seen in training, and Sundial is U for Amazon because its corpus includes in-house collections that are not itemized.
The overlapping taxi sources cover earlier years than our 2022--2024 series.
For web traffic, GIFT-Eval Pretrain contains all 145{,}063 pages of our source over the same dates, differing only in how missing values are stored, so TSFMs pretrained on it may have seen our evaluated windows.

\begin{table}[tb]
\caption{Pretraining-source overlap between each TSFM and dataset, according to the TSFM's own documentation. O: disclosed source overlap; E: disclosed exclusion or absence from an itemized corpus; U: insufficient disclosure.}
\label{tab:checkpoint-overlap}
\begin{center}
\begin{tabular}{lccccc}
\toprule
TSFM & Amazon & Favorita & M5 & NYC taxi & Web traffic \\
\midrule
Chronos-2 & E & E & E & O & O \\
Chronos-Bolt & U & U & E & U & U \\
Moirai-1.1 & U & U & U & U & U \\
Moirai-2.0 & E & O & O & O & O \\
Moirai-MoE & E & O & O & O & O \\
Sundial & U & O & E & O & O \\
TimeMoE & E & O & O & O & O \\
Timer & E & O & O & O & O \\
TimesFM-2.5 & E & O & O & O & O \\
TimesFM-3 & E & E & E & O & O \\
TiRex & E & O & E & O & O \\
Toto-2.0 & E & E & E & E & E \\
\bottomrule
\end{tabular}

\end{center}
\end{table}

Table~\ref{tab:disclosed-exclusion} groups the linear-probe contrasts of Table~\ref{tab:scoreboard} by this classification.
Disclosed exclusion does not verify that no evaluated series occurred in pretraining.
Eight of the 12 E pairs are on Amazon or M5, whereas NYC taxi and web traffic contribute only Toto-2.0, so differences between the classes do not isolate an effect of overlap.

\begin{table}[tb]
\caption{Paired linear-probe contrasts by disclosed pretraining overlap of the backbone--dataset pair (Table~\ref{tab:checkpoint-overlap}), AP and AUC skill, equal-window weighting.
Each cell gives the median over pairs and, in parentheses, how many paired 95\% intervals lie above zero, span zero, or lie below zero; the number of pairs is in the header.
All includes the five U pairs.}
\label{tab:disclosed-exclusion}
\begin{center}
\small
\resizebox{\linewidth}{!}{\begin{tabular}{lcccccc}
\toprule
 & \multicolumn{3}{c}{AP skill} & \multicolumn{3}{c}{AUC skill} \\
\cmidrule(lr){2-4}\cmidrule(lr){5-7}
Contrast & E (12) & O (13) & All (30) & E (12) & O (13) & All (30) \\
\midrule
Pretrained $-$ randomized & 0.019 (11/1/0) & 0.035 (10/0/3) & 0.027 (26/1/3) & 0.028 (10/1/1) & 0.041 (10/0/3) & 0.031 (25/1/4) \\
Probe $-$ point forecast & 0.034 (11/1/0) & 0.046 (13/0/0) & 0.034 (29/1/0) & 0.052 (11/1/0) & 0.063 (13/0/0) & 0.057 (29/1/0) \\
Probe $-$ raw logistic & 0.005 (8/2/2) & 0.035 (10/0/3) & 0.009 (22/3/5) & 0.008 (9/2/1) & 0.039 (10/0/3) & 0.015 (22/4/4) \\
\bottomrule
\end{tabular}
}
\end{center}
\end{table}

\FloatBarrier
\section{Within-window ranking metrics}
\label{app:metrics}

A ranking unit has $n$ positions, $k$ positives with $0<k<n$, and finite scores.
Sorting the distinct scores in decreasing order gives tie blocks $b=1,\dots,B$ with sizes $s_b$, positive counts $a_b$, and cumulative sums $C_b=\sum_{j\le b}s_j$ and $A_b=\sum_{j\le b}a_j$.
We verified the closed form for $\mathrm{AP}^0$ below by enumerating all 2{,}604 mixed labelings over every tie-block composition with $n\le 6$; our implementation deviates from the exact values by at most $1.1\times10^{-16}$.

\subsection{Threshold AP and its permutation expectation}

Threshold AP, $\mathrm{AP}=k^{-1}\sum_b a_bA_b/C_b$, evaluates precision at each distinct score threshold, with all tied positions entering together.
A threshold on the score cannot separate tied positions, so the ranking commits only to its threshold sets.
This convention differs from AP averaged over random tie orderings \citep{mcsherry2008ties}.
For one positive tied with one negative, threshold AP is $1/2$, whereas averaging over the two orderings gives $3/4$.

Hold the scores and $k$ fixed and place the $k$ positive labels uniformly at random among the $\binom{n}{k}$ subsets of positions.
\[
\E[\ry_i]=\frac{k}{n},\qquad \E[\ry_i\ry_j]=\frac{k(k-1)}{n(n-1)}.
\]
Writing $\ry_i\in\{0,1\}$ for the random label of position $i$, the product $a_bA_b=\sum_{i\in b}\sum_{j:\,\mathrm{block}(j)\le b}\ry_i\ry_j$ contains $s_b$ diagonal terms $\ry_i^2=\ry_i$ and $s_b(C_b-1)$ off-diagonal products, so
\[
\E[a_bA_b]=\frac{s_bk}{n}+\frac{s_b(C_b-1)k(k-1)}{n(n-1)}.
\]
By linearity, and using $\sum_b s_b=n$,
\[
\mathrm{AP}^0=\E[\mathrm{AP}]
=\sum_{b}\frac{s_b}{nC_b}\left[1+\frac{(C_b-1)(k-1)}{n-1}\right]
=\frac{k-1}{n-1}+\frac{n-k}{n(n-1)}\sum_{b}\frac{s_b}{C_b}.
\]
For a single block (a constant score), $\sum_b s_b/C_b=1$ and the expectation equals the prevalence $k/n$.
Because $k<n$, some labeling places a negative in the top tie block and has $\mathrm{AP}<1$, so $\mathrm{AP}^0<1$ and the skill is well defined.
The skill $S^{\mathrm{AP}}=(\mathrm{AP}-\mathrm{AP}^0)/(1-\mathrm{AP}^0)$ has zero conditional permutation mean, because $\mathrm{AP}^0$ is fixed given the scores and $k$.
It is identically zero for constant scores and is retained when negative.
Skills are normalized per window before averaging; we never normalize an average AP by an average reference.

The permutation law defines a normalization only.
It does not assert that observed labels are exchangeable, and no hypothesis test is based on it.

Because $1-\mathrm{AP}^0$ shrinks as $k$ approaches $n$, AP skill is bounded above by 1 but not symmetrically below; for tie-free scores with $n=64$ and $k=63$, its minimum is about $-3.0$.
Two tie-free scores on the same window share $\mathrm{AP}^0_w$, so their AP-skill difference equals their AP difference divided by $1-\mathrm{AP}^0_w$, which up-weights windows with few zeros.
Excluding such windows leaves all reported signs unchanged (Appendix~\ref{app:high-prevalence}).

\subsection{AUC skill and invariance to per-window transformations}

AUROC averages $\1_{z_i>z_j}+\tfrac12\1_{z_i=z_j}$ over the $k(n-k)$ pairs of a positive $i$ and a negative $j$, and $S^{\mathrm{AUC}}=2\,\mathrm{AUROC}-1$.
Under the permutation law above, $\E[\mathrm{AUROC}]=1/2$.

For a window $w$, let $z'_{w,h}=g_w(z_{w,h})$ for a strictly increasing function $g_w$, such as a positive affine rescaling.
Then $z'_{w,h}>z'_{w,j}$ if and only if $z_{w,h}>z_{w,j}$, and $z'_{w,h}=z'_{w,j}$ if and only if $z_{w,h}=z_{w,j}$, so pairwise order and ties are preserved.
Hence the tie blocks and their labels, AP, $\mathrm{AP}^0_w$, and every AUC pair credit are unchanged.
Eligibility depends only on labels, so every fixed weighting of window skills is invariant as well.
The argument requires the transformation to preserve order and exact ties in floating point.
A pooled ranking compares positions across windows that may be transformed differently and therefore need not be invariant.

\subsection{Pooling can reward between-window activity}
\label{app:metrics-pooling}

Consider two windows of four positions.
The first has labels 1100 and constant score 1; the second has labels 1000 and constant score 0.
Within each window, AP equals the prevalence, AUROC is $1/2$, and both skills are zero.
Pooling the eight positions into one ranking yields prevalence $3/8$, $\mathrm{AP}=11/24$, $\mathrm{AP}^0=47/112$, AP skill $1/15$, $\mathrm{AUROC}=19/30$, and AUC skill $4/15$.
The pooled ranking thus rewards the higher score of the more active window, even though neither window's score discriminates between its own positions.

\subsection{Positive AP skill under equal marginal event probabilities}
\label{app:metrics-dependent}

Fix scores $(4,3,2,1)$ and draw the labels uniformly from $\{1100, 0110, 0011, 1001\}$.
Every draw has $k=2$, and every position has marginal event probability $1/2$.
Table~\ref{tab:equal-marginals} lists AP and AUROC for each draw.

\begin{table}[htb]
\caption{Construction with equal marginal event probabilities. The scores are fixed at $(4,3,2,1)$, and each label vector is drawn with probability $1/4$, so every position has marginal event probability $1/2$. Mean AP is $11/16$, above $\mathrm{AP}^0=49/72$, whereas mean AUROC is $1/2$.}
\label{tab:equal-marginals}
\begin{center}
\footnotesize
\begin{tabular}{lccc}
\toprule
Labels & Probability & AP & AUROC \\
\midrule
1100 & $1/4$ & $1$ & $1$ \\
0110 & $1/4$ & $7/12$ & $1/2$ \\
0011 & $1/4$ & $5/12$ & $0$ \\
1001 & $1/4$ & $3/4$ & $1/2$ \\
\bottomrule
\end{tabular}
\end{center}
\end{table}

Mean AP is $11/16$, whereas $\mathrm{AP}^0$, which averages all six labelings with $k=2$, is $49/72$.
Mean AP skill is therefore $(11/16-49/72)/(1-49/72)=1/46$, while mean AUROC is $1/2$ and mean AUC skill is zero.
For fixed $k$, AUROC is affine in the sum of positive midranks, so equal marginals suffice to fix its mean.
AP instead depends on products of labels and therefore on their joint distribution.
The construction limits interpreting positive AP skill as evidence of unequal marginal event probabilities across positions.
It does not invalidate AP, does not make AUC generally preferable, and does not identify the cause of the empirical AP/AUC disagreements in Section~\ref{sec:results-references}.

\FloatBarrier
\section{Implementation details}
\label{app:provenance}

\subsection{Point forecasts and representations}
\label{app:impl-tsfm}
All TSFMs receive only the target context and forecast 64 steps.
Table~\ref{tab:point-forecast} lists the released checkpoint and the ranked point forecast of each TSFM.

\begin{table}[tb]
\caption{Released checkpoint, ranked point forecast, generation interface, and number of forward passes for the 64-step horizon. An autoregressive TSFM needs one pass when a single output step covers the horizon.}
\label{tab:point-forecast}
\begin{center}
\small
\resizebox{\textwidth}{!}{%
\begin{tabular}{llllr}
\toprule
TSFM & Checkpoint & Ranked point forecast & Generation interface & Forward passes \\
\midrule
Chronos-2 & \texttt{amazon/chronos-2} & median & direct & 1 \\
Chronos-Bolt & \texttt{amazon/chronos-bolt-base} & median & direct & 1 \\
Moirai-1.1 & \texttt{Salesforce/moirai-1.1-R-large} & mean of 20 samples & sampling & 1 \\
Moirai-2.0 & \texttt{Salesforce/moirai-2.0-R-small} & median & autoregressive & 1 \\
Moirai-MoE & \texttt{Salesforce/moirai-moe-1.0-R-base} & mean of 20 samples & autoregressive sampling & 4 \\
Sundial & \texttt{thuml/sundial-base-128m} & mean of 20 samples & sampling & 1 \\
TimeMoE & \texttt{Maple728/TimeMoE-200M} & model point output & autoregressive & 1 \\
Timer & \texttt{thuml/timer-base-84m} & model point output & autoregressive & 1 \\
TimesFM-2.5 & \texttt{google/timesfm-2.5-200m-pytorch} & median & direct & 1 \\
TimesFM-3 & \texttt{google/timesfm-3.0-pytorch} & median & direct & 1 \\
TiRex & \texttt{NX-AI/TiRex} & median & direct & 2 \\
Toto-2.0 & \texttt{Datadog/Toto-2.0-2.5B} & median & autoregressive & 1 \\
\bottomrule
\end{tabular}%
}
\end{center}
\end{table}

\paragraph{Probe selection.}
Ten of the twelve TSFMs produce the 64-step horizon in one forward pass (Table~\ref{tab:point-forecast}); Moirai-MoE and TiRex require several.
We probe six of these ten, a subset fixed before any probe was trained.
The other four, Moirai-1.1, Moirai-2.0, TimesFM-2.5, and TimesFM-3, include the strongest point forecasters on Favorita, M5, and web traffic.
Whether probes of the four omitted backbones gain as much over their own forecasts, and whether their features beat randomized copies, remains open.

\paragraph{Representation extraction.}
Except for Chronos-2, each probe reads the input of the forecast head, captured during the forecast pass (Table~\ref{tab:representation-extraction}).
For Sundial, whose flow-matching head samples, this input is the conditioning state before sampling.
For Chronos-2, the probe reads the output-patch query token returned by the model's embedding interface.
Toto-2.0 yields two vectors per window, each feeding the head for 32 horizon positions.
The six backbones run in bfloat16, and representations are converted to float32.
Each randomized backbone is built from its released configuration without loading the pretrained weights, so its parameters follow the default initialization of its source code.

\begin{table}[tb]
\caption{Forecast-aligned tensors used by the frozen-representation probes. Feature dimensions are per anchor.}
\label{tab:representation-extraction}
\begin{center}
\small
\begin{tabular}{llrrr}
\toprule
Backbone & Captured tensor & Dimension & Anchors & Positions/anchor \\
\midrule
Chronos-Bolt & output-patch-head input & 768 & 1 & 64 \\
Chronos-2 & output-patch query token & 768 & 1 & 64 \\
TimeMoE & last-token LM-head input & 768 & 1 & 64 \\
Timer & last-token LM-head input & 1{,}024 & 1 & 64 \\
Sundial & flow-head conditioning state & 768 & 1 & 64 \\
Toto-2.0 & output-head input & 2{,}048 & 2 & 32 \\
\bottomrule
\end{tabular}
\end{center}
\end{table}

\subsection{Training-free references}
\label{app:impl-references}
For a context $x_1,\dots,x_C$ and a period $p$ that divides $C$, the event-rate profile scores horizon position $h$ by the mean of $\1_{x_j>0}$ over the context positions $j\equiv h \pmod p$, which share the phase of position $C+h$; the mean profile averages $x_j$ over the same positions.
The two ramps score position $h$ by $h$ and by $-h$.
Daily datasets use $p=7$, giving four candidates, and NYC taxi uses $p\in\{24,168\}$, giving six.
On each split, the candidate with the highest equal-window AP skill on the eligible validation windows is selected.
The weekly mean profile is selected on all Favorita and M5 splits and on one web-traffic split, the decreasing ramp on the other four web-traffic splits and on two Amazon splits, the weekly event-rate profile on the other three Amazon splits, and the hour-of-day mean profile on NYC taxi.

\subsection{Probe and raw-context training}
\label{app:impl-training}
All four supervised procedures predict $y_{w,h}$ and are trained on at most 100{,}000 windows, sampled with the split seed and stratified by context event count in up to 20 quantile bins (Table~\ref{tab:readout-training}).
Raw-context and representation procedures use the same training, validation, and test windows.
Appendix~\ref{app:training-size} examines the sensitivity to the training cap.

\begin{table}[tb]
\caption{Training procedures for representation and raw-context comparisons. $D$ denotes the per-anchor representation dimension.}
\label{tab:readout-training}
\begin{center}
\small
\resizebox{\textwidth}{!}{%
\begin{tabular}{llllll}
\toprule
Procedure & Input & Standardization & Training sample & Validation selection & Output organization \\
\midrule
Raw logistic & 364/336-step context & train mean/std & 100k cap; 20 strata & L2 $\{10^{-5},\ldots,10^{1}\}$ & one linear $C\!\to\!64$ map \\
Representation linear & frozen representation & train mean/std & 100k cap; 20 strata & L2 $\{10^{-5},\ldots,10^{1}\}$ & one $D\!\to\!64$ map \\
Raw LightGBM & raw context & none & 100k cap; 20 strata & per-position early stop & 64 classifiers \\
Representation GBM & frozen representation & none & 100k cap; 20 strata & per-position early stop & 64 classifiers \\
\bottomrule
\end{tabular}%
}
\end{center}
\end{table}
The linear probe maps the frozen representation $\vh_w\in\R^{D}$ of window $w$ to position scores $\vz_w=\mW\vh_w+\vb$, with $\mW\in\R^{64\times D}$ and $\vb\in\R^{64}$; raw logistic regression uses the standardized context $\vx_w\in\R^{C}$, of length $C$, in place of $\vh_w$.
For Toto-2.0, both representation procedures share a 32-output probe, $\mW\in\R^{32\times D}$, across the two vectors of a window.
Both linear procedures select the L2 penalty by validation AP pooled over positions; the ellipses in Table~\ref{tab:readout-training} denote consecutive powers of ten.
Pooled AP serves only this selection, is applied identically to probes and raw logistic regression, and enters none of the reported scores.
LightGBM uses 31 leaves, learning rate 0.05, at most 2{,}000 rounds, and early stopping after 50 rounds without improved validation AP; a single-class validation position uses 200 fixed rounds, and a single-class training position receives a constant probability.

\FloatBarrier
\section{Complete results by dataset}
\label{app:complete}
\begin{figure}[tb]
\begin{center}
\includegraphics[width=\textwidth]{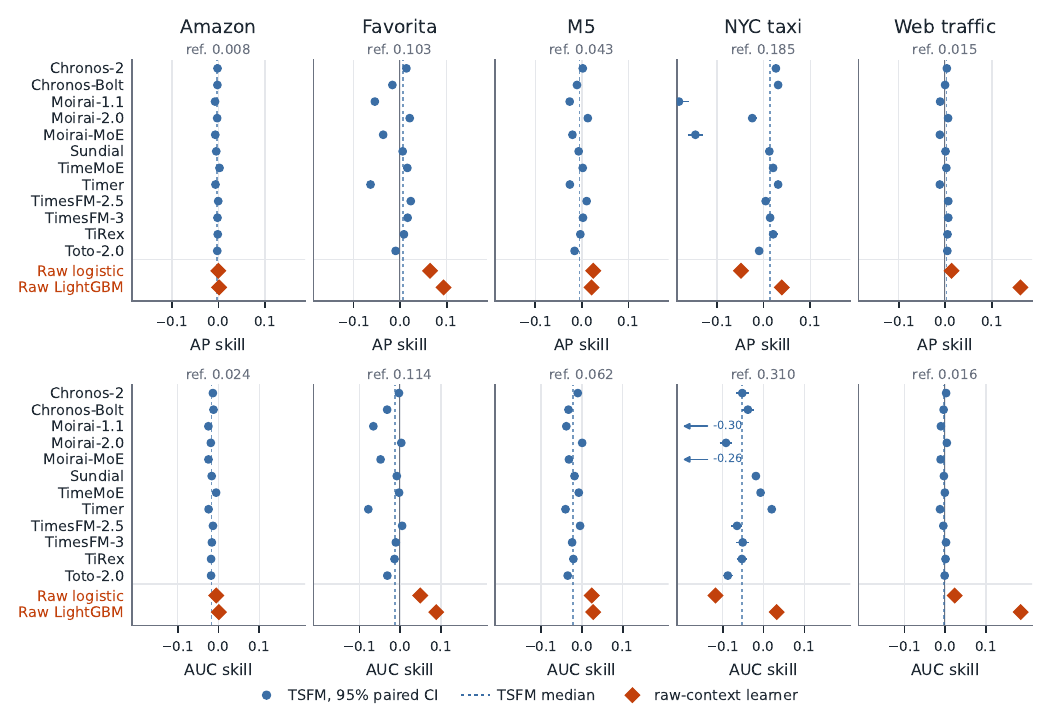}
\end{center}
\caption{Point forecasts of the 12 TSFMs and the two raw-context learners as paired differences from the training-free reference (zero line), with 95\% paired bootstrap intervals over series; equal-window weighting.
AP skill (top) and AUC skill (bottom); axes are shared within each row, and arrows mark differences beyond the axis range.
Dashed lines mark the TSFM median; panel titles give the reference's skill.}
\label{fig:absolute}
\end{figure}

Figure~\ref{fig:absolute} compares the point forecasts of all 12 TSFMs and the two raw-context learners with the training-free reference.
Tables~\ref{tab:complete-amazon}--\ref{tab:complete-web-traffic} report the underlying values per dataset together with the paired contrasts of the pretrained probes, and Table~\ref{tab:randomized} compares pretrained and randomized backbones.
All values pool the test windows of all splits and use equal-window weighting; equal-series values are in Appendix~\ref{app:weighting}.

\begin{table}[tbp]
\caption{Complete results on Amazon (22{,}600 windows, 1{,}135 series).
Top: point forecasts of the 12 TSFMs and the raw-context learners, and their difference from the training-free reference.
Bottom: pretrained linear and GBM probes of the six backbones, and their differences from the backbone's point forecast and from the raw-context learners.
Differences are row minus comparator with 95\% paired bootstrap intervals over series; equal-window weighting.
Bold marks the highest TSFM skill.}
\label{tab:complete-amazon}
\begin{center}
\scriptsize
\setlength{\tabcolsep}{3pt}
\begin{tabular}{lrr@{\ }lrr@{\ }l}
\toprule
 & \multicolumn{3}{c}{AP skill} & \multicolumn{3}{c}{AUC skill} \\
\cmidrule(lr){2-4}\cmidrule(lr){5-7}
 & Skill & \multicolumn{2}{c}{$-$ reference} & Skill & \multicolumn{2}{c}{$-$ reference} \\
\midrule
Training-free reference & 0.0078 & -- & & 0.0238 & -- & \\
\midrule
Chronos-2 & 0.0057 & $-$0.0021 & [$-$0.0042, 0.0003] & 0.0100 & $-$0.0137 & [$-$0.0203, $-$0.0072] \\
Chronos-Bolt & 0.0055 & $-$0.0023 & [$-$0.0041, $-$0.0004] & 0.0116 & $-$0.0122 & [$-$0.0192, $-$0.0050] \\
Moirai-1.1 & 0.0004 & $-$0.0074 & [$-$0.0094, $-$0.0054] & $-$0.0009 & $-$0.0247 & [$-$0.0310, $-$0.0181] \\
Moirai-2.0 & 0.0050 & $-$0.0028 & [$-$0.0045, $-$0.0011] & 0.0050 & $-$0.0188 & [$-$0.0255, $-$0.0120] \\
Moirai-MoE & 0.0009 & $-$0.0069 & [$-$0.0088, $-$0.0050] & $-$0.0007 & $-$0.0245 & [$-$0.0311, $-$0.0177] \\
Sundial & 0.0030 & $-$0.0048 & [$-$0.0068, $-$0.0028] & 0.0072 & $-$0.0166 & [$-$0.0233, $-$0.0101] \\
TimeMoE & \textbf{0.0100} & 0.0022 & [0.0000, 0.0043] & \textbf{0.0180} & $-$0.0058 & [$-$0.0122, 0.0009] \\
Timer & 0.0014 & $-$0.0064 & [$-$0.0083, $-$0.0045] & $-$0.0005 & $-$0.0243 & [$-$0.0312, $-$0.0177] \\
TimesFM-2.5 & 0.0072 & $-$0.0006 & [$-$0.0025, 0.0015] & 0.0103 & $-$0.0135 & [$-$0.0198, $-$0.0071] \\
TimesFM-3 & 0.0057 & $-$0.0022 & [$-$0.0043, 0.0000] & 0.0076 & $-$0.0162 & [$-$0.0227, $-$0.0096] \\
TiRex & 0.0062 & $-$0.0016 & [$-$0.0038, 0.0008] & 0.0056 & $-$0.0181 & [$-$0.0253, $-$0.0113] \\
Toto-2.0 & 0.0052 & $-$0.0026 & [$-$0.0046, $-$0.0005] & 0.0056 & $-$0.0182 & [$-$0.0243, $-$0.0122] \\
\midrule
Raw logistic & 0.0074 & $-$0.0004 & [$-$0.0022, 0.0017] & 0.0181 & $-$0.0057 & [$-$0.0118, 0.0004] \\
Raw LightGBM & 0.0091 & 0.0013 & [$-$0.0006, 0.0034] & 0.0246 & 0.0009 & [$-$0.0053, 0.0070] \\
\bottomrule
\end{tabular}

\vspace{1em}
\begin{tabular}{lrr@{\ }lr@{\ }lr@{\ }l}
\toprule
 & Skill & \multicolumn{2}{c}{$-$ point forecast} & \multicolumn{2}{c}{$-$ raw logistic} & \multicolumn{2}{c}{$-$ raw LightGBM} \\
\midrule
\multicolumn{8}{l}{\textit{AP skill, linear probe}} \\
Chronos-2 & 0.0130 & 0.0073 & [0.0052, 0.0094] & 0.0056 & [0.0039, 0.0073] & 0.0039 & [0.0019, 0.0059] \\
Chronos-Bolt & 0.0096 & 0.0040 & [0.0022, 0.0060] & 0.0021 & [0.0004, 0.0038] & 0.0004 & [$-$0.0016, 0.0025] \\
Sundial & 0.0093 & 0.0063 & [0.0041, 0.0083] & 0.0019 & [$-$0.0001, 0.0037] & 0.0002 & [$-$0.0019, 0.0022] \\
TimeMoE & 0.0084 & $-$0.0016 & [$-$0.0039, 0.0006] & 0.0010 & [$-$0.0010, 0.0029] & $-$0.0007 & [$-$0.0026, 0.0012] \\
Timer & 0.0064 & 0.0050 & [0.0029, 0.0071] & $-$0.0011 & [$-$0.0029, 0.0008] & $-$0.0028 & [$-$0.0048, $-$0.0007] \\
Toto-2.0 & 0.0135 & 0.0083 & [0.0066, 0.0100] & 0.0060 & [0.0038, 0.0082] & 0.0043 & [0.0022, 0.0065] \\
\midrule
\multicolumn{8}{l}{\textit{AP skill, GBM probe}} \\
Chronos-2 & 0.0044 & $-$0.0013 & [$-$0.0036, 0.0007] & $-$0.0030 & [$-$0.0050, $-$0.0011] & $-$0.0047 & [$-$0.0069, $-$0.0026] \\
Chronos-Bolt & 0.0022 & $-$0.0034 & [$-$0.0053, $-$0.0014] & $-$0.0053 & [$-$0.0073, $-$0.0032] & $-$0.0070 & [$-$0.0091, $-$0.0048] \\
Sundial & 0.0029 & $-$0.0002 & [$-$0.0023, 0.0021] & $-$0.0046 & [$-$0.0066, $-$0.0026] & $-$0.0063 & [$-$0.0085, $-$0.0041] \\
TimeMoE & 0.0015 & $-$0.0085 & [$-$0.0110, $-$0.0062] & $-$0.0060 & [$-$0.0080, $-$0.0040] & $-$0.0077 & [$-$0.0100, $-$0.0055] \\
Timer & 0.0019 & 0.0005 & [$-$0.0015, 0.0024] & $-$0.0056 & [$-$0.0077, $-$0.0036] & $-$0.0073 & [$-$0.0095, $-$0.0051] \\
Toto-2.0 & 0.0049 & $-$0.0003 & [$-$0.0024, 0.0016] & $-$0.0026 & [$-$0.0048, $-$0.0006] & $-$0.0043 & [$-$0.0065, $-$0.0022] \\
\midrule
\multicolumn{8}{l}{\textit{AUC skill, linear probe}} \\
Chronos-2 & 0.0327 & 0.0227 & [0.0170, 0.0284] & 0.0146 & [0.0097, 0.0197] & 0.0081 & [0.0025, 0.0138] \\
Chronos-Bolt & 0.0226 & 0.0110 & [0.0052, 0.0170] & 0.0045 & [$-$0.0006, 0.0094] & $-$0.0021 & [$-$0.0076, 0.0039] \\
Sundial & 0.0233 & 0.0161 & [0.0105, 0.0216] & 0.0052 & [$-$0.0005, 0.0106] & $-$0.0013 & [$-$0.0073, 0.0046] \\
TimeMoE & 0.0209 & 0.0029 & [$-$0.0031, 0.0087] & 0.0028 & [$-$0.0029, 0.0084] & $-$0.0038 & [$-$0.0093, 0.0018] \\
Timer & 0.0160 & 0.0165 & [0.0101, 0.0230] & $-$0.0022 & [$-$0.0078, 0.0035] & $-$0.0087 & [$-$0.0145, $-$0.0030] \\
Toto-2.0 & 0.0266 & 0.0211 & [0.0154, 0.0270] & 0.0085 & [0.0026, 0.0143] & 0.0020 & [$-$0.0040, 0.0078] \\
\midrule
\multicolumn{8}{l}{\textit{AUC skill, GBM probe}} \\
Chronos-2 & 0.0106 & 0.0006 & [$-$0.0055, 0.0066] & $-$0.0075 & [$-$0.0133, $-$0.0019] & $-$0.0141 & [$-$0.0201, $-$0.0080] \\
Chronos-Bolt & 0.0075 & $-$0.0040 & [$-$0.0104, 0.0026] & $-$0.0106 & [$-$0.0166, $-$0.0048] & $-$0.0171 & [$-$0.0237, $-$0.0109] \\
Sundial & 0.0067 & $-$0.0005 & [$-$0.0064, 0.0053] & $-$0.0115 & [$-$0.0173, $-$0.0057] & $-$0.0180 & [$-$0.0242, $-$0.0118] \\
TimeMoE & 0.0047 & $-$0.0133 & [$-$0.0199, $-$0.0069] & $-$0.0134 & [$-$0.0198, $-$0.0074] & $-$0.0200 & [$-$0.0266, $-$0.0133] \\
Timer & 0.0072 & 0.0077 & [0.0016, 0.0140] & $-$0.0109 & [$-$0.0169, $-$0.0048] & $-$0.0174 & [$-$0.0236, $-$0.0112] \\
Toto-2.0 & 0.0093 & 0.0037 & [$-$0.0019, 0.0095] & $-$0.0088 & [$-$0.0150, $-$0.0029] & $-$0.0153 & [$-$0.0219, $-$0.0089] \\
\bottomrule
\end{tabular}

\end{center}
\end{table}

\begin{table}[tbp]
\caption{Complete results on Favorita (321{,}747 windows, 26{,}427 series); layout as in Table~\ref{tab:complete-amazon}.}
\label{tab:complete-favorita}
\begin{center}
\scriptsize
\setlength{\tabcolsep}{3pt}
\begin{tabular}{lrr@{\ }lrr@{\ }l}
\toprule
 & \multicolumn{3}{c}{AP skill} & \multicolumn{3}{c}{AUC skill} \\
\cmidrule(lr){2-4}\cmidrule(lr){5-7}
 & Skill & \multicolumn{2}{c}{$-$ reference} & Skill & \multicolumn{2}{c}{$-$ reference} \\
\midrule
Training-free reference & 0.1026 & -- & & 0.1136 & -- & \\
\midrule
Chronos-2 & 0.1160 & 0.0133 & [0.0123, 0.0144] & 0.1106 & $-$0.0030 & [$-$0.0043, $-$0.0018] \\
Chronos-Bolt & 0.0859 & $-$0.0167 & [$-$0.0180, $-$0.0155] & 0.0816 & $-$0.0320 & [$-$0.0334, $-$0.0306] \\
Moirai-1.1 & 0.0482 & $-$0.0544 & [$-$0.0558, $-$0.0530] & 0.0478 & $-$0.0658 & [$-$0.0673, $-$0.0643] \\
Moirai-2.0 & 0.1231 & 0.0204 & [0.0195, 0.0215] & 0.1163 & 0.0027 & [0.0014, 0.0039] \\
Moirai-MoE & 0.0661 & $-$0.0365 & [$-$0.0378, $-$0.0352] & 0.0655 & $-$0.0481 & [$-$0.0497, $-$0.0465] \\
Sundial & 0.1080 & 0.0054 & [0.0042, 0.0066] & 0.1052 & $-$0.0084 & [$-$0.0098, $-$0.0069] \\
TimeMoE & 0.1180 & 0.0153 & [0.0139, 0.0167] & 0.1108 & $-$0.0027 & [$-$0.0044, $-$0.0011] \\
Timer & 0.0390 & $-$0.0637 & [$-$0.0657, $-$0.0617] & 0.0355 & $-$0.0781 & [$-$0.0801, $-$0.0761] \\
TimesFM-2.5 & \textbf{0.1254} & 0.0228 & [0.0217, 0.0238] & \textbf{0.1184} & 0.0048 & [0.0036, 0.0060] \\
TimesFM-3 & 0.1186 & 0.0160 & [0.0150, 0.0170] & 0.1029 & $-$0.0107 & [$-$0.0119, $-$0.0095] \\
TiRex & 0.1107 & 0.0081 & [0.0070, 0.0091] & 0.0997 & $-$0.0139 & [$-$0.0153, $-$0.0126] \\
Toto-2.0 & 0.0928 & $-$0.0098 & [$-$0.0110, $-$0.0087] & 0.0821 & $-$0.0315 & [$-$0.0329, $-$0.0301] \\
\midrule
Raw logistic & 0.1668 & 0.0642 & [0.0621, 0.0662] & 0.1628 & 0.0492 & [0.0470, 0.0515] \\
Raw LightGBM & 0.1957 & 0.0931 & [0.0911, 0.0952] & 0.2020 & 0.0884 & [0.0862, 0.0906] \\
\bottomrule
\end{tabular}

\vspace{1em}
\begin{tabular}{lrr@{\ }lr@{\ }lr@{\ }l}
\toprule
 & Skill & \multicolumn{2}{c}{$-$ point forecast} & \multicolumn{2}{c}{$-$ raw logistic} & \multicolumn{2}{c}{$-$ raw LightGBM} \\
\midrule
\multicolumn{8}{l}{\textit{AP skill, linear probe}} \\
Chronos-2 & 0.1932 & 0.0772 & [0.0755, 0.0791] & 0.0264 & [0.0255, 0.0274] & $-$0.0025 & [$-$0.0035, $-$0.0015] \\
Chronos-Bolt & 0.1763 & 0.0904 & [0.0884, 0.0923] & 0.0095 & [0.0084, 0.0106] & $-$0.0194 & [$-$0.0204, $-$0.0184] \\
Sundial & 0.1942 & 0.0862 & [0.0845, 0.0879] & 0.0274 & [0.0264, 0.0284] & $-$0.0015 & [$-$0.0025, $-$0.0005] \\
TimeMoE & 0.1743 & 0.0563 & [0.0547, 0.0578] & 0.0075 & [0.0065, 0.0085] & $-$0.0214 & [$-$0.0225, $-$0.0204] \\
Timer & 0.1347 & 0.0958 & [0.0937, 0.0977] & $-$0.0321 & [$-$0.0333, $-$0.0309] & $-$0.0610 & [$-$0.0622, $-$0.0597] \\
Toto-2.0 & 0.1647 & 0.0720 & [0.0702, 0.0737] & $-$0.0020 & [$-$0.0031, $-$0.0009] & $-$0.0309 & [$-$0.0320, $-$0.0299] \\
\midrule
\multicolumn{8}{l}{\textit{AP skill, GBM probe}} \\
Chronos-2 & 0.1811 & 0.0651 & [0.0633, 0.0668] & 0.0143 & [0.0132, 0.0153] & $-$0.0146 & [$-$0.0156, $-$0.0137] \\
Chronos-Bolt & 0.1668 & 0.0809 & [0.0790, 0.0828] & 0.0000 & [$-$0.0010, 0.0011] & $-$0.0289 & [$-$0.0299, $-$0.0279] \\
Sundial & 0.1779 & 0.0699 & [0.0683, 0.0716] & 0.0111 & [0.0101, 0.0122] & $-$0.0178 & [$-$0.0187, $-$0.0168] \\
TimeMoE & 0.1624 & 0.0444 & [0.0429, 0.0459] & $-$0.0044 & [$-$0.0055, $-$0.0033] & $-$0.0333 & [$-$0.0343, $-$0.0323] \\
Timer & 0.1094 & 0.0704 & [0.0684, 0.0723] & $-$0.0574 & [$-$0.0587, $-$0.0560] & $-$0.0863 & [$-$0.0877, $-$0.0849] \\
Toto-2.0 & 0.1541 & 0.0613 & [0.0595, 0.0630] & $-$0.0127 & [$-$0.0138, $-$0.0116] & $-$0.0416 & [$-$0.0427, $-$0.0406] \\
\midrule
\multicolumn{8}{l}{\textit{AUC skill, linear probe}} \\
Chronos-2 & 0.1938 & 0.0832 & [0.0814, 0.0850] & 0.0310 & [0.0301, 0.0321] & $-$0.0082 & [$-$0.0091, $-$0.0072] \\
Chronos-Bolt & 0.1760 & 0.0944 & [0.0926, 0.0962] & 0.0132 & [0.0122, 0.0143] & $-$0.0260 & [$-$0.0270, $-$0.0250] \\
Sundial & 0.1936 & 0.0884 & [0.0867, 0.0901] & 0.0308 & [0.0297, 0.0318] & $-$0.0084 & [$-$0.0094, $-$0.0075] \\
TimeMoE & 0.1744 & 0.0636 & [0.0619, 0.0652] & 0.0117 & [0.0106, 0.0127] & $-$0.0276 & [$-$0.0286, $-$0.0265] \\
Timer & 0.1417 & 0.1062 & [0.1043, 0.1080] & $-$0.0211 & [$-$0.0223, $-$0.0200] & $-$0.0603 & [$-$0.0616, $-$0.0592] \\
Toto-2.0 & 0.1661 & 0.0840 & [0.0821, 0.0858] & 0.0033 & [0.0021, 0.0044] & $-$0.0359 & [$-$0.0370, $-$0.0349] \\
\midrule
\multicolumn{8}{l}{\textit{AUC skill, GBM probe}} \\
Chronos-2 & 0.1854 & 0.0748 & [0.0732, 0.0765] & 0.0227 & [0.0215, 0.0238] & $-$0.0166 & [$-$0.0175, $-$0.0156] \\
Chronos-Bolt & 0.1699 & 0.0883 & [0.0865, 0.0900] & 0.0071 & [0.0061, 0.0082] & $-$0.0321 & [$-$0.0331, $-$0.0311] \\
Sundial & 0.1796 & 0.0743 & [0.0727, 0.0760] & 0.0168 & [0.0157, 0.0179] & $-$0.0224 & [$-$0.0234, $-$0.0215] \\
TimeMoE & 0.1639 & 0.0531 & [0.0515, 0.0547] & 0.0012 & [0.0002, 0.0023] & $-$0.0381 & [$-$0.0390, $-$0.0370] \\
Timer & 0.1198 & 0.0843 & [0.0825, 0.0861] & $-$0.0430 & [$-$0.0443, $-$0.0417] & $-$0.0822 & [$-$0.0836, $-$0.0810] \\
Toto-2.0 & 0.1584 & 0.0763 & [0.0745, 0.0781] & $-$0.0043 & [$-$0.0055, $-$0.0033] & $-$0.0436 & [$-$0.0446, $-$0.0425] \\
\bottomrule
\end{tabular}

\end{center}
\end{table}

\begin{table}[tbp]
\caption{Complete results on M5 (120{,}063 windows, 5{,}979 series); layout as in Table~\ref{tab:complete-amazon}.}
\label{tab:complete-m5}
\begin{center}
\scriptsize
\setlength{\tabcolsep}{3pt}
\begin{tabular}{lrr@{\ }lrr@{\ }l}
\toprule
 & \multicolumn{3}{c}{AP skill} & \multicolumn{3}{c}{AUC skill} \\
\cmidrule(lr){2-4}\cmidrule(lr){5-7}
 & Skill & \multicolumn{2}{c}{$-$ reference} & Skill & \multicolumn{2}{c}{$-$ reference} \\
\midrule
Training-free reference & 0.0434 & -- & & 0.0619 & -- & \\
\midrule
Chronos-2 & 0.0453 & 0.0019 & [0.0006, 0.0032] & 0.0518 & $-$0.0101 & [$-$0.0121, $-$0.0080] \\
Chronos-Bolt & 0.0329 & $-$0.0105 & [$-$0.0119, $-$0.0092] & 0.0290 & $-$0.0329 & [$-$0.0350, $-$0.0307] \\
Moirai-1.1 & 0.0173 & $-$0.0262 & [$-$0.0277, $-$0.0247] & 0.0237 & $-$0.0382 & [$-$0.0404, $-$0.0361] \\
Moirai-2.0 & \textbf{0.0561} & 0.0127 & [0.0114, 0.0140] & \textbf{0.0626} & 0.0007 & [$-$0.0015, 0.0028] \\
Moirai-MoE & 0.0232 & $-$0.0203 & [$-$0.0217, $-$0.0190] & 0.0301 & $-$0.0318 & [$-$0.0340, $-$0.0296] \\
Sundial & 0.0365 & $-$0.0070 & [$-$0.0083, $-$0.0057] & 0.0438 & $-$0.0181 & [$-$0.0203, $-$0.0159] \\
TimeMoE & 0.0452 & 0.0018 & [0.0001, 0.0034] & 0.0543 & $-$0.0076 & [$-$0.0100, $-$0.0052] \\
Timer & 0.0176 & $-$0.0258 & [$-$0.0277, $-$0.0239] & 0.0217 & $-$0.0402 & [$-$0.0427, $-$0.0377] \\
TimesFM-2.5 & 0.0536 & 0.0102 & [0.0090, 0.0115] & 0.0574 & $-$0.0045 & [$-$0.0066, $-$0.0024] \\
TimesFM-3 & 0.0458 & 0.0023 & [0.0012, 0.0035] & 0.0379 & $-$0.0240 & [$-$0.0260, $-$0.0220] \\
TiRex & 0.0401 & $-$0.0033 & [$-$0.0047, $-$0.0020] & 0.0410 & $-$0.0209 & [$-$0.0230, $-$0.0188] \\
Toto-2.0 & 0.0277 & $-$0.0157 & [$-$0.0171, $-$0.0143] & 0.0270 & $-$0.0348 & [$-$0.0370, $-$0.0327] \\
\midrule
Raw logistic & 0.0678 & 0.0243 & [0.0228, 0.0258] & 0.0861 & 0.0243 & [0.0221, 0.0264] \\
Raw LightGBM & 0.0642 & 0.0207 & [0.0192, 0.0223] & 0.0896 & 0.0277 & [0.0253, 0.0301] \\
\bottomrule
\end{tabular}

\vspace{1em}
\begin{tabular}{lrr@{\ }lr@{\ }lr@{\ }l}
\toprule
 & Skill & \multicolumn{2}{c}{$-$ point forecast} & \multicolumn{2}{c}{$-$ raw logistic} & \multicolumn{2}{c}{$-$ raw LightGBM} \\
\midrule
\multicolumn{8}{l}{\textit{AP skill, linear probe}} \\
Chronos-2 & 0.0762 & 0.0309 & [0.0296, 0.0321] & 0.0084 & [0.0074, 0.0094] & 0.0120 & [0.0111, 0.0130] \\
Chronos-Bolt & 0.0708 & 0.0379 & [0.0365, 0.0393] & 0.0030 & [0.0020, 0.0041] & 0.0066 & [0.0056, 0.0077] \\
Sundial & 0.0714 & 0.0349 & [0.0334, 0.0363] & 0.0036 & [0.0026, 0.0046] & 0.0072 & [0.0062, 0.0082] \\
TimeMoE & 0.0608 & 0.0155 & [0.0144, 0.0168] & $-$0.0070 & [$-$0.0081, $-$0.0059] & $-$0.0034 & [$-$0.0044, $-$0.0024] \\
Timer & 0.0462 & 0.0286 & [0.0270, 0.0300] & $-$0.0216 & [$-$0.0229, $-$0.0202] & $-$0.0180 & [$-$0.0192, $-$0.0167] \\
Toto-2.0 & 0.0599 & 0.0322 & [0.0309, 0.0335] & $-$0.0079 & [$-$0.0090, $-$0.0067] & $-$0.0043 & [$-$0.0053, $-$0.0031] \\
\midrule
\multicolumn{8}{l}{\textit{AP skill, GBM probe}} \\
Chronos-2 & 0.0595 & 0.0142 & [0.0129, 0.0154] & $-$0.0083 & [$-$0.0094, $-$0.0073] & $-$0.0047 & [$-$0.0057, $-$0.0036] \\
Chronos-Bolt & 0.0551 & 0.0222 & [0.0209, 0.0235] & $-$0.0127 & [$-$0.0137, $-$0.0115] & $-$0.0091 & [$-$0.0101, $-$0.0080] \\
Sundial & 0.0549 & 0.0184 & [0.0171, 0.0198] & $-$0.0129 & [$-$0.0140, $-$0.0117] & $-$0.0093 & [$-$0.0103, $-$0.0082] \\
TimeMoE & 0.0482 & 0.0030 & [0.0017, 0.0041] & $-$0.0196 & [$-$0.0208, $-$0.0183] & $-$0.0160 & [$-$0.0172, $-$0.0149] \\
Timer & 0.0337 & 0.0161 & [0.0147, 0.0175] & $-$0.0340 & [$-$0.0356, $-$0.0325] & $-$0.0304 & [$-$0.0318, $-$0.0290] \\
Toto-2.0 & 0.0448 & 0.0171 & [0.0158, 0.0185] & $-$0.0229 & [$-$0.0242, $-$0.0216] & $-$0.0193 & [$-$0.0206, $-$0.0181] \\
\midrule
\multicolumn{8}{l}{\textit{AUC skill, linear probe}} \\
Chronos-2 & 0.1016 & 0.0498 & [0.0479, 0.0517] & 0.0155 & [0.0140, 0.0168] & 0.0120 & [0.0107, 0.0132] \\
Chronos-Bolt & 0.0920 & 0.0631 & [0.0609, 0.0653] & 0.0059 & [0.0045, 0.0073] & 0.0024 & [0.0011, 0.0037] \\
Sundial & 0.0939 & 0.0501 & [0.0481, 0.0520] & 0.0078 & [0.0062, 0.0093] & 0.0043 & [0.0029, 0.0056] \\
TimeMoE & 0.0823 & 0.0280 & [0.0262, 0.0298] & $-$0.0038 & [$-$0.0053, $-$0.0021] & $-$0.0073 & [$-$0.0087, $-$0.0060] \\
Timer & 0.0666 & 0.0450 & [0.0428, 0.0469] & $-$0.0195 & [$-$0.0212, $-$0.0177] & $-$0.0230 & [$-$0.0245, $-$0.0215] \\
Toto-2.0 & 0.0803 & 0.0533 & [0.0512, 0.0553] & $-$0.0058 & [$-$0.0075, $-$0.0042] & $-$0.0093 & [$-$0.0108, $-$0.0079] \\
\midrule
\multicolumn{8}{l}{\textit{AUC skill, GBM probe}} \\
Chronos-2 & 0.0815 & 0.0298 & [0.0278, 0.0316] & $-$0.0046 & [$-$0.0062, $-$0.0031] & $-$0.0081 & [$-$0.0096, $-$0.0068] \\
Chronos-Bolt & 0.0766 & 0.0476 & [0.0454, 0.0497] & $-$0.0095 & [$-$0.0111, $-$0.0078] & $-$0.0130 & [$-$0.0144, $-$0.0116] \\
Sundial & 0.0756 & 0.0318 & [0.0298, 0.0337] & $-$0.0105 & [$-$0.0122, $-$0.0088] & $-$0.0140 & [$-$0.0154, $-$0.0126] \\
TimeMoE & 0.0652 & 0.0109 & [0.0091, 0.0128] & $-$0.0209 & [$-$0.0227, $-$0.0193] & $-$0.0244 & [$-$0.0260, $-$0.0230] \\
Timer & 0.0517 & 0.0301 & [0.0280, 0.0320] & $-$0.0344 & [$-$0.0363, $-$0.0324] & $-$0.0379 & [$-$0.0396, $-$0.0362] \\
Toto-2.0 & 0.0636 & 0.0366 & [0.0345, 0.0387] & $-$0.0225 & [$-$0.0243, $-$0.0207] & $-$0.0260 & [$-$0.0276, $-$0.0244] \\
\bottomrule
\end{tabular}

\end{center}
\end{table}

\begin{table}[tbp]
\caption{Complete results on NYC taxi (19{,}207 windows, 164 series, one calendar split); layout as in Table~\ref{tab:complete-amazon}.}
\label{tab:complete-nyc-taxi}
\begin{center}
\scriptsize
\setlength{\tabcolsep}{3pt}
\begin{tabular}{lrr@{\ }lrr@{\ }l}
\toprule
 & \multicolumn{3}{c}{AP skill} & \multicolumn{3}{c}{AUC skill} \\
\cmidrule(lr){2-4}\cmidrule(lr){5-7}
 & Skill & \multicolumn{2}{c}{$-$ reference} & Skill & \multicolumn{2}{c}{$-$ reference} \\
\midrule
Training-free reference & 0.1853 & -- & & 0.3095 & -- & \\
\midrule
Chronos-2 & 0.2117 & 0.0265 & [0.0171, 0.0363] & 0.2572 & $-$0.0523 & [$-$0.0672, $-$0.0366] \\
Chronos-Bolt & \textbf{0.2164} & 0.0311 & [0.0226, 0.0395] & 0.2710 & $-$0.0385 & [$-$0.0541, $-$0.0224] \\
Moirai-1.1 & 0.0035 & $-$0.1817 & [$-$0.2058, $-$0.1607] & 0.0051 & $-$0.3044 & [$-$0.3292, $-$0.2799] \\
Moirai-2.0 & 0.1609 & $-$0.0244 & [$-$0.0339, $-$0.0150] & 0.2173 & $-$0.0922 & [$-$0.1076, $-$0.0768] \\
Moirai-MoE & 0.0388 & $-$0.1464 & [$-$0.1623, $-$0.1310] & 0.0460 & $-$0.2635 & [$-$0.2834, $-$0.2442] \\
Sundial & 0.1976 & 0.0123 & [0.0063, 0.0182] & 0.2906 & $-$0.0189 & [$-$0.0260, $-$0.0124] \\
TimeMoE & 0.2057 & 0.0204 & [0.0146, 0.0265] & 0.3020 & $-$0.0075 & [$-$0.0149, $-$0.0002] \\
Timer & \textbf{0.2164} & 0.0311 & [0.0267, 0.0360] & \textbf{0.3294} & 0.0199 & [0.0144, 0.0251] \\
TimesFM-2.5 & 0.1897 & 0.0045 & [$-$0.0041, 0.0132] & 0.2444 & $-$0.0651 & [$-$0.0789, $-$0.0517] \\
TimesFM-3 & 0.1992 & 0.0140 & [0.0053, 0.0234] & 0.2583 & $-$0.0512 & [$-$0.0667, $-$0.0357] \\
TiRex & 0.2059 & 0.0206 & [0.0130, 0.0299] & 0.2564 & $-$0.0531 & [$-$0.0660, $-$0.0400] \\
Toto-2.0 & 0.1757 & $-$0.0095 & [$-$0.0167, $-$0.0019] & 0.2217 & $-$0.0878 & [$-$0.0996, $-$0.0760] \\
\midrule
Raw logistic & 0.1367 & $-$0.0485 & [$-$0.0544, $-$0.0422] & 0.1915 & $-$0.1180 & [$-$0.1273, $-$0.1083] \\
Raw LightGBM & 0.2243 & 0.0390 & [0.0338, 0.0453] & 0.3417 & 0.0322 & [0.0269, 0.0382] \\
\bottomrule
\end{tabular}

\vspace{1em}
\begin{tabular}{lrr@{\ }lr@{\ }lr@{\ }l}
\toprule
 & Skill & \multicolumn{2}{c}{$-$ point forecast} & \multicolumn{2}{c}{$-$ raw logistic} & \multicolumn{2}{c}{$-$ raw LightGBM} \\
\midrule
\multicolumn{8}{l}{\textit{AP skill, linear probe}} \\
Chronos-2 & 0.2237 & 0.0120 & [0.0058, 0.0180] & 0.0870 & [0.0806, 0.0935] & $-$0.0006 & [$-$0.0030, 0.0017] \\
Chronos-Bolt & 0.2346 & 0.0182 & [0.0139, 0.0227] & 0.0978 & [0.0897, 0.1060] & 0.0103 & [0.0067, 0.0137] \\
Sundial & 0.2234 & 0.0258 & [0.0228, 0.0286] & 0.0867 & [0.0796, 0.0936] & $-$0.0009 & [$-$0.0044, 0.0022] \\
TimeMoE & 0.2181 & 0.0124 & [0.0100, 0.0149] & 0.0814 & [0.0745, 0.0885] & $-$0.0061 & [$-$0.0096, $-$0.0029] \\
Timer & 0.2229 & 0.0065 & [0.0045, 0.0085] & 0.0862 & [0.0803, 0.0922] & $-$0.0014 & [$-$0.0036, 0.0008] \\
Toto-2.0 & 0.2319 & 0.0562 & [0.0519, 0.0607] & 0.0952 & [0.0878, 0.1029] & 0.0076 & [0.0047, 0.0105] \\
\midrule
\multicolumn{8}{l}{\textit{AP skill, GBM probe}} \\
Chronos-2 & 0.2009 & $-$0.0109 & [$-$0.0190, $-$0.0031] & 0.0641 & [0.0584, 0.0700] & $-$0.0234 & [$-$0.0271, $-$0.0201] \\
Chronos-Bolt & 0.2216 & 0.0052 & [$-$0.0002, 0.0106] & 0.0849 & [0.0776, 0.0922] & $-$0.0027 & [$-$0.0055, $-$0.0001] \\
Sundial & 0.2026 & 0.0050 & [0.0014, 0.0085] & 0.0659 & [0.0594, 0.0723] & $-$0.0217 & [$-$0.0260, $-$0.0180] \\
TimeMoE & 0.1843 & $-$0.0214 & [$-$0.0252, $-$0.0178] & 0.0476 & [0.0420, 0.0531] & $-$0.0400 & [$-$0.0441, $-$0.0362] \\
Timer & 0.2056 & $-$0.0107 & [$-$0.0130, $-$0.0085] & 0.0689 & [0.0632, 0.0749] & $-$0.0186 & [$-$0.0217, $-$0.0155] \\
Toto-2.0 & 0.2158 & 0.0401 & [0.0344, 0.0460] & 0.0791 & [0.0715, 0.0869] & $-$0.0085 & [$-$0.0123, $-$0.0050] \\
\midrule
\multicolumn{8}{l}{\textit{AUC skill, linear probe}} \\
Chronos-2 & 0.3451 & 0.0879 & [0.0726, 0.1029] & 0.1536 & [0.1443, 0.1631] & 0.0034 & [$-$0.0002, 0.0068] \\
Chronos-Bolt & 0.3516 & 0.0806 & [0.0662, 0.0956] & 0.1601 & [0.1502, 0.1702] & 0.0099 & [0.0047, 0.0148] \\
Sundial & 0.3462 & 0.0556 & [0.0497, 0.0617] & 0.1547 & [0.1448, 0.1640] & 0.0046 & [$-$0.0002, 0.0086] \\
TimeMoE & 0.3348 & 0.0327 & [0.0271, 0.0383] & 0.1433 & [0.1332, 0.1531] & $-$0.0069 & [$-$0.0114, $-$0.0028] \\
Timer & 0.3483 & 0.0188 & [0.0150, 0.0230] & 0.1568 & [0.1471, 0.1658] & 0.0066 & [0.0029, 0.0100] \\
Toto-2.0 & 0.3503 & 0.1285 & [0.1166, 0.1401] & 0.1588 & [0.1490, 0.1683] & 0.0086 & [0.0047, 0.0123] \\
\midrule
\multicolumn{8}{l}{\textit{AUC skill, GBM probe}} \\
Chronos-2 & 0.3126 & 0.0554 & [0.0405, 0.0704] & 0.1211 & [0.1096, 0.1316] & $-$0.0291 & [$-$0.0351, $-$0.0239] \\
Chronos-Bolt & 0.3336 & 0.0626 & [0.0477, 0.0786] & 0.1421 & [0.1311, 0.1532] & $-$0.0081 & [$-$0.0133, $-$0.0030] \\
Sundial & 0.3103 & 0.0196 & [0.0140, 0.0254] & 0.1188 & [0.1074, 0.1296] & $-$0.0314 & [$-$0.0377, $-$0.0258] \\
TimeMoE & 0.2792 & $-$0.0228 & [$-$0.0281, $-$0.0177] & 0.0877 & [0.0775, 0.0977] & $-$0.0625 & [$-$0.0678, $-$0.0574] \\
Timer & 0.3269 & $-$0.0025 & [$-$0.0069, 0.0021] & 0.1354 & [0.1250, 0.1456] & $-$0.0147 & [$-$0.0197, $-$0.0101] \\
Toto-2.0 & 0.3215 & 0.0998 & [0.0881, 0.1119] & 0.1300 & [0.1181, 0.1419] & $-$0.0202 & [$-$0.0266, $-$0.0140] \\
\bottomrule
\end{tabular}

\end{center}
\end{table}

\begin{table}[tbp]
\caption{Complete results on web traffic (41{,}816 windows, 1{,}597 series); layout as in Table~\ref{tab:complete-amazon}.}
\label{tab:complete-web-traffic}
\begin{center}
\scriptsize
\setlength{\tabcolsep}{3pt}
\begin{tabular}{lrr@{\ }lrr@{\ }l}
\toprule
 & \multicolumn{3}{c}{AP skill} & \multicolumn{3}{c}{AUC skill} \\
\cmidrule(lr){2-4}\cmidrule(lr){5-7}
 & Skill & \multicolumn{2}{c}{$-$ reference} & Skill & \multicolumn{2}{c}{$-$ reference} \\
\midrule
Training-free reference & 0.0154 & -- & & 0.0155 & -- & \\
\midrule
Chronos-2 & 0.0187 & 0.0033 & [0.0002, 0.0064] & 0.0175 & 0.0020 & [$-$0.0020, 0.0062] \\
Chronos-Bolt & 0.0149 & $-$0.0005 & [$-$0.0036, 0.0027] & 0.0113 & $-$0.0042 & [$-$0.0085, $-$0.0002] \\
Moirai-1.1 & 0.0043 & $-$0.0111 & [$-$0.0139, $-$0.0085] & 0.0046 & $-$0.0109 & [$-$0.0148, $-$0.0070] \\
Moirai-2.0 & 0.0214 & 0.0059 & [0.0029, 0.0091] & \textbf{0.0192} & 0.0037 & [$-$0.0004, 0.0077] \\
Moirai-MoE & 0.0037 & $-$0.0117 & [$-$0.0146, $-$0.0089] & 0.0039 & $-$0.0116 & [$-$0.0155, $-$0.0077] \\
Sundial & 0.0159 & 0.0005 & [$-$0.0028, 0.0039] & 0.0120 & $-$0.0035 & [$-$0.0078, 0.0007] \\
TimeMoE & 0.0176 & 0.0022 & [$-$0.0008, 0.0054] & 0.0144 & $-$0.0011 & [$-$0.0056, 0.0029] \\
Timer & 0.0036 & $-$0.0118 & [$-$0.0147, $-$0.0089] & 0.0026 & $-$0.0129 & [$-$0.0172, $-$0.0089] \\
TimesFM-2.5 & \textbf{0.0218} & 0.0064 & [0.0032, 0.0095] & 0.0106 & $-$0.0049 & [$-$0.0092, $-$0.0006] \\
TimesFM-3 & \textbf{0.0218} & 0.0063 & [0.0031, 0.0098] & 0.0173 & 0.0018 & [$-$0.0023, 0.0063] \\
TiRex & 0.0201 & 0.0047 & [0.0016, 0.0078] & 0.0160 & 0.0005 & [$-$0.0040, 0.0048] \\
Toto-2.0 & 0.0197 & 0.0043 & [0.0014, 0.0074] & 0.0139 & $-$0.0016 & [$-$0.0054, 0.0022] \\
\midrule
Raw logistic & 0.0288 & 0.0134 & [0.0106, 0.0162] & 0.0384 & 0.0229 & [0.0190, 0.0267] \\
Raw LightGBM & 0.1768 & 0.1614 & [0.1512, 0.1723] & 0.2003 & 0.1848 & [0.1735, 0.1967] \\
\bottomrule
\end{tabular}

\vspace{1em}
\begin{tabular}{lrr@{\ }lr@{\ }lr@{\ }l}
\toprule
 & Skill & \multicolumn{2}{c}{$-$ point forecast} & \multicolumn{2}{c}{$-$ raw logistic} & \multicolumn{2}{c}{$-$ raw LightGBM} \\
\midrule
\multicolumn{8}{l}{\textit{AP skill, linear probe}} \\
Chronos-2 & 0.0667 & 0.0480 & [0.0440, 0.0518] & 0.0379 & [0.0338, 0.0418] & $-$0.1101 & [$-$0.1188, $-$0.1009] \\
Chronos-Bolt & 0.0570 & 0.0421 & [0.0386, 0.0455] & 0.0282 & [0.0246, 0.0318] & $-$0.1198 & [$-$0.1290, $-$0.1104] \\
Sundial & 0.0725 & 0.0566 & [0.0522, 0.0608] & 0.0437 & [0.0393, 0.0480] & $-$0.1043 & [$-$0.1129, $-$0.0957] \\
TimeMoE & 0.0635 & 0.0459 & [0.0417, 0.0501] & 0.0347 & [0.0305, 0.0386] & $-$0.1133 & [$-$0.1219, $-$0.1042] \\
Timer & 0.0594 & 0.0558 & [0.0516, 0.0600] & 0.0305 & [0.0268, 0.0341] & $-$0.1175 & [$-$0.1261, $-$0.1084] \\
Toto-2.0 & 0.0606 & 0.0409 & [0.0369, 0.0448] & 0.0318 & [0.0276, 0.0359] & $-$0.1162 & [$-$0.1251, $-$0.1070] \\
\midrule
\multicolumn{8}{l}{\textit{AP skill, GBM probe}} \\
Chronos-2 & 0.1016 & 0.0829 & [0.0760, 0.0898] & 0.0728 & [0.0660, 0.0792] & $-$0.0752 & [$-$0.0814, $-$0.0692] \\
Chronos-Bolt & 0.1020 & 0.0871 & [0.0800, 0.0939] & 0.0732 & [0.0663, 0.0796] & $-$0.0748 & [$-$0.0809, $-$0.0688] \\
Sundial & 0.1180 & 0.1021 & [0.0941, 0.1098] & 0.0891 & [0.0814, 0.0965] & $-$0.0588 & [$-$0.0645, $-$0.0533] \\
TimeMoE & 0.0974 & 0.0797 & [0.0729, 0.0865] & 0.0686 & [0.0621, 0.0748] & $-$0.0794 & [$-$0.0856, $-$0.0731] \\
Timer & 0.1060 & 0.1024 & [0.0940, 0.1104] & 0.0772 & [0.0695, 0.0844] & $-$0.0708 & [$-$0.0760, $-$0.0656] \\
Toto-2.0 & 0.0895 & 0.0698 & [0.0634, 0.0757] & 0.0607 & [0.0544, 0.0666] & $-$0.0873 & [$-$0.0945, $-$0.0803] \\
\midrule
\multicolumn{8}{l}{\textit{AUC skill, linear probe}} \\
Chronos-2 & 0.0797 & 0.0622 & [0.0569, 0.0674] & 0.0413 & [0.0370, 0.0456] & $-$0.1207 & [$-$0.1299, $-$0.1109] \\
Chronos-Bolt & 0.0697 & 0.0584 & [0.0528, 0.0640] & 0.0313 & [0.0273, 0.0352] & $-$0.1306 & [$-$0.1405, $-$0.1205] \\
Sundial & 0.0866 & 0.0745 & [0.0684, 0.0801] & 0.0482 & [0.0434, 0.0530] & $-$0.1138 & [$-$0.1228, $-$0.1046] \\
TimeMoE & 0.0769 & 0.0625 & [0.0573, 0.0682] & 0.0385 & [0.0337, 0.0432] & $-$0.1234 & [$-$0.1325, $-$0.1138] \\
Timer & 0.0740 & 0.0714 & [0.0659, 0.0768] & 0.0356 & [0.0310, 0.0399] & $-$0.1264 & [$-$0.1355, $-$0.1170] \\
Toto-2.0 & 0.0700 & 0.0560 & [0.0504, 0.0613] & 0.0316 & [0.0273, 0.0357] & $-$0.1304 & [$-$0.1401, $-$0.1204] \\
\midrule
\multicolumn{8}{l}{\textit{AUC skill, GBM probe}} \\
Chronos-2 & 0.1183 & 0.1008 & [0.0928, 0.1089] & 0.0799 & [0.0725, 0.0870] & $-$0.0820 & [$-$0.0889, $-$0.0753] \\
Chronos-Bolt & 0.1207 & 0.1094 & [0.1005, 0.1183] & 0.0823 & [0.0745, 0.0897] & $-$0.0796 & [$-$0.0863, $-$0.0729] \\
Sundial & 0.1354 & 0.1234 & [0.1137, 0.1328] & 0.0970 & [0.0884, 0.1053] & $-$0.0649 & [$-$0.0710, $-$0.0586] \\
TimeMoE & 0.1141 & 0.0997 & [0.0915, 0.1079] & 0.0757 & [0.0681, 0.0829] & $-$0.0863 & [$-$0.0931, $-$0.0791] \\
Timer & 0.1258 & 0.1232 & [0.1135, 0.1325] & 0.0874 & [0.0788, 0.0959] & $-$0.0745 & [$-$0.0801, $-$0.0688] \\
Toto-2.0 & 0.1038 & 0.0898 & [0.0818, 0.0975] & 0.0654 & [0.0586, 0.0719] & $-$0.0965 & [$-$0.1042, $-$0.0888] \\
\bottomrule
\end{tabular}

\end{center}
\end{table}

\begin{table}[tbp]
\caption{Pretrained versus randomized backbones under the same probe, equal-window weighting.
Pretr.\ and Rand.\ give the skill of the pretrained and the randomized probe; the difference is pretrained minus randomized, with 95\% paired bootstrap intervals over series.
The linear probe covers all six backbones on all five datasets, the GBM probe Chronos-2 and Toto-2.0 on the four daily datasets.
NYC taxi has one fitted run per initialization.}
\label{tab:randomized}
\begin{center}
\resizebox{\textwidth}{!}{\begin{tabular}{lllrrr@{\ }lrrr@{\ }l}
\toprule
 & & & \multicolumn{4}{c}{AP skill} & \multicolumn{4}{c}{AUC skill} \\
\cmidrule(lr){4-7}\cmidrule(lr){8-11}
Backbone & Dataset & Probe & Pretr. & Rand. & \multicolumn{2}{c}{Difference [95\% CI]} & Pretr. & Rand. & \multicolumn{2}{c}{Difference [95\% CI]} \\
\midrule
Chronos-2 & Amazon & linear & 0.0130 & 0.0022 & 0.0108 & [0.0086, 0.0130] & 0.0327 & 0.0035 & 0.0292 & [0.0231, 0.0356] \\
Chronos-2 & Amazon & GBM & 0.0044 & $-$0.0001 & 0.0045 & [0.0026, 0.0066] & 0.0106 & 0.0012 & 0.0094 & [0.0037, 0.0153] \\
Chronos-2 & Favorita & linear & 0.1932 & 0.0801 & 0.1131 & [0.1113, 0.1149] & 0.1938 & 0.0817 & 0.1121 & [0.1105, 0.1136] \\
Chronos-2 & Favorita & GBM & 0.1811 & 0.0661 & 0.1150 & [0.1132, 0.1168] & 0.1854 & 0.0692 & 0.1163 & [0.1146, 0.1178] \\
Chronos-2 & M5 & linear & 0.0762 & 0.0148 & 0.0614 & [0.0593, 0.0635] & 0.1016 & 0.0224 & 0.0792 & [0.0769, 0.0814] \\
Chronos-2 & M5 & GBM & 0.0595 & 0.0079 & 0.0516 & [0.0496, 0.0536] & 0.0815 & 0.0114 & 0.0701 & [0.0677, 0.0724] \\
Chronos-2 & NYC taxi & linear & 0.2237 & 0.0442 & 0.1796 & [0.1568, 0.2026] & 0.3451 & 0.0787 & 0.2664 & [0.2429, 0.2904] \\
Chronos-2 & Web traffic & linear & 0.0667 & 0.0313 & 0.0354 & [0.0321, 0.0386] & 0.0797 & 0.0459 & 0.0338 & [0.0302, 0.0373] \\
Chronos-2 & Web traffic & GBM & 0.1016 & 0.0604 & 0.0412 & [0.0373, 0.0452] & 0.1183 & 0.0798 & 0.0385 & [0.0344, 0.0427] \\
\midrule
Chronos-Bolt & Amazon & linear & 0.0096 & 0.0015 & 0.0081 & [0.0062, 0.0101] & 0.0226 & 0.0075 & 0.0151 & [0.0099, 0.0206] \\
Chronos-Bolt & Favorita & linear & 0.1763 & 0.0691 & 0.1072 & [0.1054, 0.1088] & 0.1760 & 0.0698 & 0.1062 & [0.1046, 0.1078] \\
Chronos-Bolt & M5 & linear & 0.0708 & 0.0105 & 0.0603 & [0.0582, 0.0624] & 0.0920 & 0.0144 & 0.0777 & [0.0752, 0.0801] \\
Chronos-Bolt & NYC taxi & linear & 0.2346 & 0.0284 & 0.2062 & [0.1810, 0.2313] & 0.3516 & 0.0505 & 0.3012 & [0.2772, 0.3250] \\
Chronos-Bolt & Web traffic & linear & 0.0570 & 0.0321 & 0.0249 & [0.0218, 0.0281] & 0.0697 & 0.0466 & 0.0231 & [0.0195, 0.0266] \\
\midrule
Sundial & Amazon & linear & 0.0093 & 0.0043 & 0.0050 & [0.0032, 0.0069] & 0.0233 & 0.0133 & 0.0100 & [0.0043, 0.0153] \\
Sundial & Favorita & linear & 0.1942 & 0.1407 & 0.0535 & [0.0522, 0.0547] & 0.1936 & 0.1430 & 0.0506 & [0.0494, 0.0517] \\
Sundial & M5 & linear & 0.0714 & 0.0437 & 0.0277 & [0.0264, 0.0289] & 0.0939 & 0.0609 & 0.0329 & [0.0314, 0.0345] \\
Sundial & NYC taxi & linear & 0.2234 & 0.1600 & 0.0634 & [0.0573, 0.0696] & 0.3462 & 0.2390 & 0.1072 & [0.1008, 0.1133] \\
Sundial & Web traffic & linear & 0.0725 & 0.0553 & 0.0172 & [0.0145, 0.0199] & 0.0866 & 0.0725 & 0.0141 & [0.0106, 0.0175] \\
\midrule
TimeMoE & Amazon & linear & 0.0084 & 0.0040 & 0.0044 & [0.0023, 0.0065] & 0.0209 & 0.0092 & 0.0117 & [0.0058, 0.0176] \\
TimeMoE & Favorita & linear & 0.1743 & 0.0881 & 0.0862 & [0.0846, 0.0878] & 0.1744 & 0.0917 & 0.0828 & [0.0814, 0.0843] \\
TimeMoE & M5 & linear & 0.0608 & 0.0262 & 0.0345 & [0.0328, 0.0362] & 0.0823 & 0.0374 & 0.0449 & [0.0430, 0.0470] \\
TimeMoE & NYC taxi & linear & 0.2181 & 0.1033 & 0.1148 & [0.1011, 0.1289] & 0.3348 & 0.1563 & 0.1785 & [0.1663, 0.1911] \\
TimeMoE & Web traffic & linear & 0.0635 & 0.0361 & 0.0274 & [0.0242, 0.0307] & 0.0769 & 0.0482 & 0.0287 & [0.0246, 0.0328] \\
\midrule
Timer & Amazon & linear & 0.0064 & 0.0073 & $-$0.0010 & [$-$0.0029, 0.0009] & 0.0160 & 0.0218 & $-$0.0058 & [$-$0.0113, $-$0.0003] \\
Timer & Favorita & linear & 0.1347 & 0.1723 & $-$0.0376 & [$-$0.0388, $-$0.0364] & 0.1417 & 0.1751 & $-$0.0335 & [$-$0.0345, $-$0.0324] \\
Timer & M5 & linear & 0.0462 & 0.0562 & $-$0.0100 & [$-$0.0112, $-$0.0087] & 0.0666 & 0.0777 & $-$0.0111 & [$-$0.0127, $-$0.0095] \\
Timer & NYC taxi & linear & 0.2229 & 0.2013 & 0.0216 & [0.0190, 0.0242] & 0.3483 & 0.3076 & 0.0407 & [0.0367, 0.0447] \\
Timer & Web traffic & linear & 0.0594 & 0.0674 & $-$0.0081 & [$-$0.0105, $-$0.0055] & 0.0740 & 0.0849 & $-$0.0109 & [$-$0.0142, $-$0.0078] \\
\midrule
Toto-2.0 & Amazon & linear & 0.0135 & 0.0097 & 0.0038 & [0.0015, 0.0059] & 0.0266 & 0.0250 & 0.0016 & [$-$0.0040, 0.0069] \\
Toto-2.0 & Amazon & GBM & 0.0049 & 0.0049 & $-$0.0001 & [$-$0.0021, 0.0020] & 0.0093 & 0.0155 & $-$0.0062 & [$-$0.0124, $-$0.0002] \\
Toto-2.0 & Favorita & linear & 0.1647 & 0.1385 & 0.0262 & [0.0250, 0.0275] & 0.1661 & 0.1399 & 0.0262 & [0.0249, 0.0275] \\
Toto-2.0 & Favorita & GBM & 0.1541 & 0.1326 & 0.0214 & [0.0202, 0.0227] & 0.1584 & 0.1382 & 0.0202 & [0.0190, 0.0214] \\
Toto-2.0 & M5 & linear & 0.0599 & 0.0486 & 0.0113 & [0.0101, 0.0125] & 0.0803 & 0.0664 & 0.0139 & [0.0123, 0.0154] \\
Toto-2.0 & M5 & GBM & 0.0448 & 0.0405 & 0.0044 & [0.0031, 0.0057] & 0.0636 & 0.0592 & 0.0045 & [0.0026, 0.0062] \\
Toto-2.0 & NYC taxi & linear & 0.2319 & 0.1591 & 0.0728 & [0.0654, 0.0797] & 0.3503 & 0.2429 & 0.1073 & [0.1007, 0.1136] \\
Toto-2.0 & Web traffic & linear & 0.0606 & 0.0516 & 0.0090 & [0.0062, 0.0120] & 0.0700 & 0.0624 & 0.0076 & [0.0041, 0.0112] \\
Toto-2.0 & Web traffic & GBM & 0.0895 & 0.1227 & $-$0.0332 & [$-$0.0386, $-$0.0274] & 0.1038 & 0.1446 & $-$0.0408 & [$-$0.0468, $-$0.0344] \\
\bottomrule
\end{tabular}
}
\end{center}
\end{table}

\FloatBarrier

\subsection{AP and AUC skill of the probed backbones}
\label{app:full-figures}
\begin{figure}[tb]
\begin{center}
\includegraphics[width=0.9\textwidth]{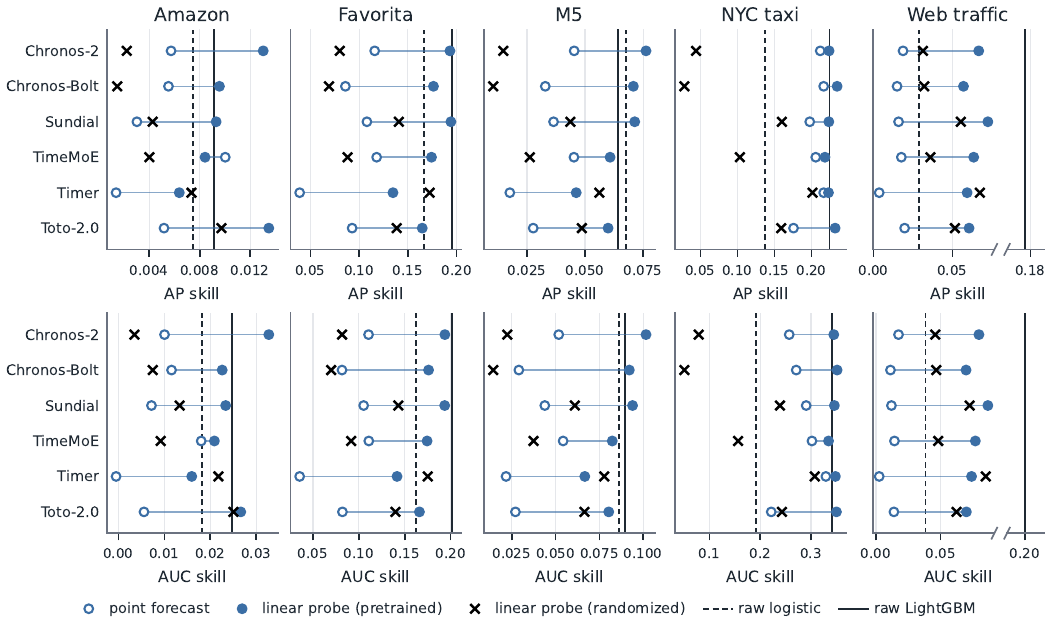}
\end{center}
\caption{AP skill (top) and AUC skill (bottom) of the six probed backbones on the five datasets, equal-window weighting. The top row repeats Figure~\ref{fig:representation}. Axes differ by dataset; the web-traffic axis is broken. Points are marginal estimates, so their overlap is not a test; per-backbone intervals are in Tables~\ref{tab:complete-amazon}--\ref{tab:complete-web-traffic}.}
\label{fig:representation-full}
\end{figure}

\begin{figure}[tb]
\begin{center}
\includegraphics[width=0.9\textwidth]{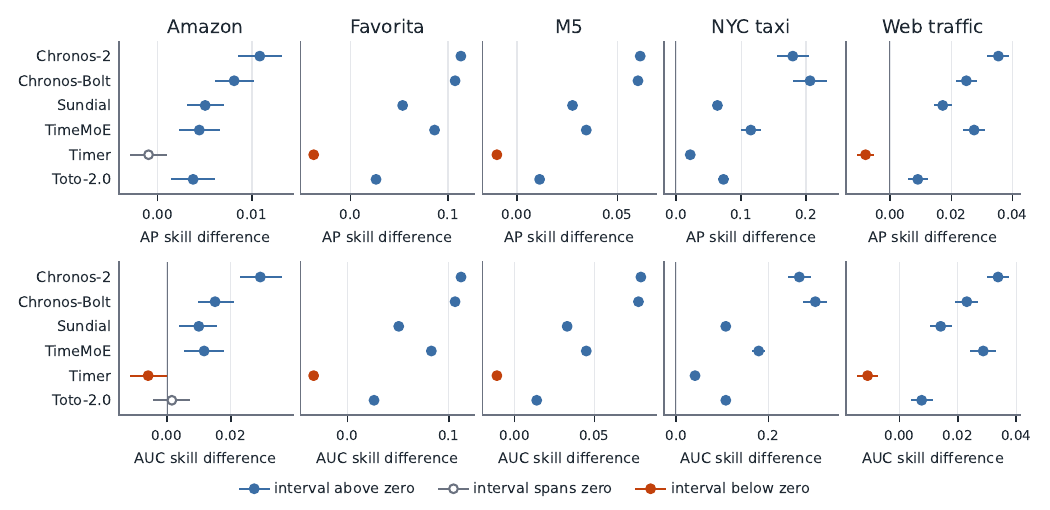}
\end{center}
\caption{Pretrained-minus-randomized difference in AP skill (top) and AUC skill (bottom) under the linear probe, with pointwise 95\% paired bootstrap intervals over series; equal-window weighting. The top row repeats Figure~\ref{fig:randomized}. Axes differ by dataset and always include zero. NYC taxi has one fitted run per initialization. Values are in Table~\ref{tab:randomized}.}
\label{fig:randomized-full}
\end{figure}

Figures~\ref{fig:representation-full} and~\ref{fig:randomized-full} repeat Figures~\ref{fig:representation} and~\ref{fig:randomized} with both metrics.
Their top rows show AP skill, as in the main text, and their bottom rows show AUC skill.

\subsection{Events among the eight highest-ranked positions}
\label{app:precision}
AP and AUC skill are zero in expectation for a random ordering and one for a perfect ranking.
For a more concrete scale, Table~\ref{tab:precision} reports how many events each method places among the eight highest-ranked of the 64 positions, that is, $8\times$ precision@8, with tied positions at the cutoff contributing their expected number of events.
A random ordering places $8k_w/64$ events there on average, and an oracle $\min(8,k_w)$.
The values are conditional on the scored windows, which contain both events and zeros; this is not known in advance, so they are not expected yields in deployment.
Web-traffic labels include missing values stored as zeros (Appendix~\ref{app:data-zeros}).

\begin{table}[tb]
\caption{Expected number of events among the eight highest-ranked positions of a window, averaged over the scored windows with equal weight.
Best point forecast: the highest value among the 12 TSFMs on each dataset, a descriptive test-set maximum.
Probes are the pretrained linear probes.}
\label{tab:precision}
\begin{center}
\small
\begin{tabular}{lrrrrr}
\toprule
Method & Amazon & Favorita & M5 & NYC taxi & Web traffic \\
\midrule
Random ordering & 1.00 & 4.34 & 2.87 & 2.11 & 2.75 \\
Training-free reference & 1.07 & 4.80 & 3.18 & 3.37 & 2.78 \\
Best point forecast & 1.06 & 4.79 & 3.16 & 3.54 & 2.86 \\
Raw logistic & 1.06 & 4.89 & 3.24 & 3.13 & 2.89 \\
Raw LightGBM & 1.07 & 5.01 & 3.22 & 3.51 & 3.65 \\
\addlinespace
Probe: Chronos-2 & 1.09 & 4.98 & 3.27 & 3.50 & 3.10 \\
Probe: Chronos-Bolt & 1.07 & 4.89 & 3.22 & 3.58 & 3.04 \\
Probe: Sundial & 1.07 & 4.97 & 3.23 & 3.49 & 3.13 \\
Probe: TimeMoE & 1.06 & 4.89 & 3.17 & 3.46 & 3.07 \\
Probe: Timer & 1.04 & 4.75 & 3.10 & 3.49 & 3.05 \\
Probe: Toto-2.0 & 1.09 & 4.86 & 3.17 & 3.55 & 3.06 \\
\addlinespace
Oracle & 5.20 & 7.62 & 7.37 & 7.04 & 7.05 \\
\bottomrule
\end{tabular}

\end{center}
\end{table}

\FloatBarrier

\section{Robustness of the comparisons}
\label{app:probe-robustness}

\subsection{Equal-window and equal-series averaging}
\label{app:weighting}
Let $\sW$ be the set of evaluated windows of a dataset, $\sG$ the set of global series with at least one window in $\sW$, and $\sW_g\subseteq \sW$ the windows of series $g$, pooled over all splits in which $g$ is a test series.
The two averages of per-window skills $S_w$ are
\[
\bar S_{\mathrm{window}}=\frac{1}{|\sW|}\sum_{w\in \sW}S_w,
\qquad
\bar S_{\mathrm{series}}=\frac{1}{|\sG|}\sum_{g\in \sG}\bar S_g,
\qquad
\bar S_g=\frac{1}{|\sW_g|}\sum_{w\in \sW_g}S_w .
\]
Equal-window averaging weights each series by its number of evaluated windows; equal-series averaging weights every series equally.
The windows, reference selections, and bootstrap clusters are identical under both.
The two averages differ by
$\bar S_{\mathrm{window}}-\bar S_{\mathrm{series}}=\Cov_{\sG}(|\sW_g|,\bar S_g)/\overline{|\sW_g|}$,
where $\overline{|\sW_g|}=|\sW|/|\sG|$ and
$\Cov_{\sG}(a,b)=|\sG|^{-1}\sum_{g\in \sG}(a_g-\bar a)(b_g-\bar b)$,
so they diverge when skill varies with the number of evaluated windows of a series.

\paragraph{Point forecasts.}
Table~\ref{tab:weighting} reports both averages for the comparison with the training-free reference.
On web traffic, the averaging unit changes the sign of this comparison under both metrics.
The reference's AP skill rises from 0.015 to 0.078 and its AUC skill from 0.016 to 0.089, so that even the best TSFM trails it by 0.039 AP skill and 0.064 AUC skill under equal-series averaging.
For TimesFM-2.5, which has the highest AP skill under both weightings, the AP difference moves from $+0.0064$ $[0.0032, 0.0095]$ to $-0.0395$ $[-0.0500, -0.0291]$ on the same 41{,}816 windows and 1{,}597 series.
The reversal is concentrated in 145 windows whose target begins with a run of events and is zero thereafter, as are all later targets of the series in that split; the decreasing horizon ramp, selected on four of the five web-traffic splits, ranks such targets perfectly, and most of these windows belong to series with one or two evaluated windows.
Excluding them, together with five windows of the mirrored pattern (0.4\% of all windows), gives an equal-series TimesFM-2.5 difference of $-0.0021$ $[-0.0085, 0.0041]$.
Raw-context LightGBM's difference from the reference remains large under both averages (AP skill 0.161 and 0.195).

On the other four datasets, the median and maximum TSFM differences keep their sign under both weightings, except that the largest AUC difference on Amazon moves from $-0.006$ to $+0.002$.
Raw-context LightGBM's differences from the reference keep their sign everywhere, and all four Amazon intervals include zero.

\begin{table}[tb]
\caption{Point forecasts under equal-window and equal-series averaging. For each dataset, metric, and weighting: the selected reference's skill, the median and maximum TSFM-minus-reference difference over the 12 TSFMs, and raw-context LightGBM's difference from the reference with its 95\% paired bootstrap interval over series. Windows and series are identical under both weightings.}
\label{tab:weighting}
\begin{center}
\small
\begin{tabular}{llrrrl}
\toprule
Dataset & Weighting & Reference & Median $-$ ref. & Max $-$ ref. & \makecell{Raw LightGBM $-$ ref.\\{}[95\% CI]} \\
\midrule
\multicolumn{6}{l}{\textit{AP skill}} \\
Amazon & window & 0.0078 & $-$0.0025 & 0.0022 & 0.0013 [$-$0.0006, 0.0034] \\
 & series & 0.0070 & $-$0.0019 & 0.0042 & 0.0016 [$-$0.0013, 0.0047] \\
Favorita & window & 0.1026 & 0.0067 & 0.0228 & 0.0931 [0.0911, 0.0952] \\
 & series & 0.0923 & 0.0069 & 0.0217 & 0.0981 [0.0953, 0.1008] \\
M5 & window & 0.0434 & $-$0.0052 & 0.0127 & 0.0207 [0.0192, 0.0223] \\
 & series & 0.0438 & $-$0.0039 & 0.0139 & 0.0232 [0.0213, 0.0251] \\
NYC taxi & window & 0.1853 & 0.0132 & 0.0311 & 0.0390 [0.0338, 0.0453] \\
 & series & 0.1659 & 0.0112 & 0.0291 & 0.0347 [0.0291, 0.0405] \\
Web traffic & window & 0.0154 & 0.0028 & 0.0064 & 0.1614 [0.1512, 0.1723] \\
 & series & 0.0783 & $-$0.0504 & $-$0.0395 & 0.1951 [0.1815, 0.2083] \\
\midrule
\multicolumn{6}{l}{\textit{AUC skill}} \\
Amazon & window & 0.0238 & $-$0.0174 & $-$0.0058 & 0.0009 [$-$0.0053, 0.0070] \\
 & series & 0.0186 & $-$0.0139 & 0.0016 & 0.0047 [$-$0.0048, 0.0148] \\
Favorita & window & 0.1136 & $-$0.0123 & 0.0048 & 0.0884 [0.0862, 0.0906] \\
 & series & 0.1026 & $-$0.0112 & 0.0047 & 0.0910 [0.0882, 0.0936] \\
M5 & window & 0.0619 & $-$0.0224 & 0.0007 & 0.0277 [0.0253, 0.0301] \\
 & series & 0.0616 & $-$0.0218 & 0.0024 & 0.0316 [0.0290, 0.0341] \\
NYC taxi & window & 0.3095 & $-$0.0527 & 0.0199 & 0.0322 [0.0269, 0.0382] \\
 & series & 0.2797 & $-$0.0522 & 0.0184 & 0.0316 [0.0177, 0.0425] \\
Web traffic & window & 0.0155 & $-$0.0025 & 0.0037 & 0.1848 [0.1735, 0.1967] \\
 & series & 0.0894 & $-$0.0832 & $-$0.0639 & 0.2114 [0.1968, 0.2256] \\
\bottomrule
\end{tabular}

\end{center}
\end{table}

\paragraph{Probe contrasts.}
Table~\ref{tab:readout-weighting} repeats the probe comparisons of Section~\ref{sec:results-representation} under equal-series weighting.
The direction of the linear-versus-GBM difference and the interval class of every web-traffic probe contrast are unchanged.
Outside Amazon, 12 of 288 probe contrasts change interval class or point-estimate sign, 9 of them on NYC taxi; on Amazon, 23 of 72 do.
No interval moves from one side of zero to the other.

\begin{table}[tb]
\caption{Probe sensitivity to the readout and the averaging unit. Top: range over the six backbones of the linear-minus-GBM probe difference; all 120 paired intervals (five datasets, six backbones, two metrics, two weightings) exclude zero. Bottom: every probe contrast outside Amazon whose interval class (above, spanning, or below zero) or point-estimate sign differs between equal-window and equal-series weighting. Such changes affect 23 of 72 contrasts on Amazon, 2 on Favorita, 1 on M5, 9 on NYC taxi, and none on web traffic; no interval moves from one side of zero to the other.}
\label{tab:readout-weighting}
\begin{center}
\small
\begin{tabular}{lllll}
\toprule
 & \multicolumn{2}{c}{AP skill, linear $-$ GBM} & \multicolumn{2}{c}{AUC skill, linear $-$ GBM} \\
\cmidrule(lr){2-3}\cmidrule(lr){4-5}
Dataset & equal-window & equal-series & equal-window & equal-series \\
\midrule
Amazon & 0.005 to 0.009 & 0.005 to 0.007 & 0.009 to 0.022 & 0.010 to 0.020 \\
Favorita & 0.009 to 0.025 & 0.010 to 0.024 & 0.006 to 0.022 & 0.005 to 0.019 \\
M5 & 0.012 to 0.017 & 0.012 to 0.017 & 0.015 to 0.020 & 0.015 to 0.020 \\
NYC taxi & 0.013 to 0.034 & 0.013 to 0.031 & 0.018 to 0.056 & 0.026 to 0.056 \\
Web traffic & $-$0.047 to $-$0.029 & $-$0.065 to $-$0.045 & $-$0.052 to $-$0.034 & $-$0.066 to $-$0.043 \\
\bottomrule
\end{tabular}

\vspace{0.8em}
\resizebox{\linewidth}{!}{%
\begin{tabular}{lllr@{\ }lr@{\ }l}
\toprule
Backbone & Contrast & Metric & \multicolumn{2}{c}{Equal-window [95\% CI]} & \multicolumn{2}{c}{Equal-series [95\% CI]} \\
\midrule
\multicolumn{7}{l}{\textit{Favorita}} \\
Chronos-Bolt & GBM $-$ raw logistic & AP & 0.0000 & [$-$0.0010, 0.0011] & $-$0.0011 & [$-$0.0028, 0.0006] \\
Sundial & linear $-$ raw LightGBM & AP & $-$0.0015 & [$-$0.0025, $-$0.0005] & 0.0004 & [$-$0.0011, 0.0020] \\
\midrule
\multicolumn{7}{l}{\textit{M5}} \\
Chronos-Bolt & linear $-$ raw LightGBM & AUC & 0.0024 & [0.0011, 0.0037] & 0.0008 & [$-$0.0008, 0.0023] \\
\midrule
\multicolumn{7}{l}{\textit{NYC taxi}} \\
Chronos-Bolt & GBM $-$ raw LightGBM & AP & $-$0.0027 & [$-$0.0055, $-$0.0001] & $-$0.0027 & [$-$0.0057, 0.0003] \\
Chronos-Bolt & GBM $-$ raw LightGBM & AUC & $-$0.0081 & [$-$0.0133, $-$0.0030] & $-$0.0132 & [$-$0.0262, 0.0014] \\
Chronos-Bolt & linear $-$ raw LightGBM & AUC & 0.0099 & [0.0047, 0.0148] & 0.0153 & [$-$0.0005, 0.0382] \\
TimeMoE & linear $-$ raw LightGBM & AUC & $-$0.0069 & [$-$0.0114, $-$0.0028] & $-$0.0074 & [$-$0.0161, 0.0022] \\
Timer & GBM $-$ raw LightGBM & AUC & $-$0.0147 & [$-$0.0197, $-$0.0101] & $-$0.0139 & [$-$0.0271, 0.0004] \\
Timer & linear $-$ raw LightGBM & AUC & 0.0066 & [0.0029, 0.0100] & 0.0134 & [$-$0.0030, 0.0375] \\
Toto-2.0 & GBM $-$ raw LightGBM & AP & $-$0.0085 & [$-$0.0123, $-$0.0050] & $-$0.0061 & [$-$0.0120, 0.0021] \\
Toto-2.0 & GBM $-$ raw LightGBM & AUC & $-$0.0202 & [$-$0.0266, $-$0.0140] & $-$0.0212 & [$-$0.0420, 0.0086] \\
Toto-2.0 & linear $-$ raw LightGBM & AUC & 0.0086 & [0.0047, 0.0123] & 0.0047 & [$-$0.0025, 0.0118] \\
\bottomrule
\end{tabular}%
}

\end{center}
\end{table}

\paragraph{Pretrained versus randomized backbones.}
Under equal-series weighting, the pretrained linear probe exceeds the randomized one with an interval above zero in 25 of 30 comparisons in AP skill and 24 of 30 in AUC skill, against 26 and 25 under equal-window weighting.
All changes are on Amazon, except for Timer on web traffic.
There, the difference changes sign in AP skill, from $-0.0081$ $[-0.0105, -0.0055]$ to $+0.0069$ $[0.0016, 0.0128]$, and its AUC-skill interval spans zero.
On Favorita and M5, Timer's randomized probe remains ahead under both metrics, with intervals below zero.
Among the 16 GBM-probe contrasts, five change interval class, all to spanning zero, including both of Toto-2.0's negative web-traffic differences.

\subsection{Split stability}
\label{app:split-stability}
The pooled intervals condition on the fitted predictors.
To assess refitting, we count, for each daily dataset, backbone, comparator, and metric, how many of the five seeded splits give a contrast with the sign of the pooled estimate (Table~\ref{tab:split-signs}); NYC taxi has a single split.
All five splits agree in 169 of the 192 contrasts.
M5 and web traffic agree throughout, and 21 of the 23 exceptions are on Amazon, where every pooled skill lies below 0.014 in AP and 0.033 in AUC.
The two exceptions on Favorita, Toto-2.0 versus raw logistic regression and Chronos-2 versus raw LightGBM in AP skill, each have one split of opposite sign.
Split agreement closely tracks the pooled intervals: of the 174 contrasts whose interval excludes zero, 168 agree in all five splits, and of the 18 whose interval spans zero, 17 do not.
Each split assigns 90\% of the same series to training, so any two splits share at least 80\% of all series as training series; each split also draws its own randomized initialization.
The counts therefore measure robustness to re-splitting and refitting, not independent replication.

\begin{table}[tb]
\caption{Split sign stability of the linear-probe contrasts on the four daily datasets, equal-window weighting. Each cell gives the number of the six backbones whose contrast has the pooled sign in all five seeded splits, for AP skill / AUC skill. NYC taxi has a single split.}
\label{tab:split-signs}
\begin{center}
\small
\begin{tabular}{lcccc}
\toprule
Contrast & Amazon & Favorita & M5 & Web traffic \\
\midrule
Probe $-$ point forecast   & 6 / 5 & 6 / 6 & 6 / 6 & 6 / 6 \\
Probe $-$ raw logistic     & 2 / 2 & 5 / 6 & 6 / 6 & 6 / 6 \\
Probe $-$ raw LightGBM     & 3 / 1 & 5 / 6 & 6 / 6 & 6 / 6 \\
Pretrained $-$ randomized  & 5 / 3 & 6 / 6 & 6 / 6 & 6 / 6 \\
\bottomrule
\end{tabular}
\end{center}
\end{table}

\subsection{Windows with few zeros}
\label{app:high-prevalence}
Windows with at least 58 events among 64 positions make up at most 14\% of the evaluated windows of any dataset, but under the TimesFM-2.5 forecast they carry up to 63\% of the total weight $\sum_w 1/(1-\mathrm{AP}^0_w)$ (Favorita).
Excluding them changes the sign of no AP-skill value in Table~\ref{tab:output-summary}, no AP-skill median in Table~\ref{tab:scoreboard}, and none of the 120 linear-probe contrasts in Tables~\ref{tab:complete-amazon}--\ref{tab:complete-web-traffic}, and every contrast interval that excluded zero still does.
Estimates move by at most 0.019 on Favorita and by at most 0.004 on the other datasets.
The same sign and interval results hold when excluding windows with at least 52 events.

\subsection{Web-traffic windows without missing values}
\label{app:web-missing}
Most web-traffic zero labels are missing page views (Appendix~\ref{app:data-zeros}).
We therefore re-score the original predictions on three subsets of the evaluated windows, without refitting any model (Table~\ref{tab:web-missing}).
Subset (a) keeps windows with no missing day in context or target, so all of their labels are observed.
Subsets (b) and (c) restrict only the context, to no missing day or to fewer than 5\% missing days, so their targets can still contain missing values stored as zeros.
Each subset keeps at most 15\% of the windows and leans towards series with more recorded activity: the median retained series has page views on 46--49\% of its days, against 32\% over all evaluated series.

Under (a), raw LightGBM stays ahead of all six linear and all six GBM probes, by a median of 0.039 and 0.049 AP skill, against 0.115 and 0.075 on all windows.
The other comparisons have weaker support.
The linear probe exceeds raw logistic regression with an interval above zero for two of six backbones, and pretrained features exceed randomized ones for three of six in AP skill and two in AUC skill; the remaining intervals span zero.
The intervals of both the best and the median TSFM against the reference span zero under (a), and both lie below zero under (b) and (c).
The subsets differ from the full dataset in both missingness and composition, so the smaller gaps do not measure how much missing values change the full-dataset estimates.
All models were trained with missing values stored as zeros, so this check concerns the evaluation labels only.

\begin{table}[tb]
\caption{Web-traffic contrasts on all evaluated windows and on subsets filtered by missing days, equal-window weighting.
(a): no missing day in context or target; (b): no missing day in context; (c): fewer than 5\% missing context days.
Targets in (b) and (c) can contain missing values stored as zeros.
Each cell gives the median over the six backbones and, in parentheses, how many paired 95\% intervals lie above zero, span zero, or lie below zero.
The TSFM rows are single comparisons; the best TSFM is a descriptive maximum selected on each subset.
Original predictions are re-scored without refitting.}
\label{tab:web-missing}
\begin{center}
\small
\resizebox{\linewidth}{!}{\begin{tabular}{lrrrr}
\toprule
 & Unfiltered & (a) Context and target & (b) Context & (c) Context $<$5\% \\
Windows / series & 41{,}816 / 1{,}597 & 1{,}907 / 210 & 2{,}406 / 445 & 6{,}208 / 673 \\
\midrule
\multicolumn{5}{l}{\emph{AP skill}} \\
LightGBM $-$ linear probe & 0.115 (6/0/0) & 0.039 (6/0/0) & 0.047 (6/0/0) & 0.041 (6/0/0) \\
LightGBM $-$ GBM probe & 0.075 (6/0/0) & 0.049 (6/0/0) & 0.058 (6/0/0) & 0.045 (6/0/0) \\
Linear probe $-$ raw logistic & 0.033 (6/0/0) & 0.018 (2/4/0) & 0.023 (4/2/0) & 0.020 (6/0/0) \\
Pretrained $-$ randomized & 0.021 (5/0/1) & 0.013 (3/3/0) & 0.052 (5/1/0) & 0.028 (5/1/0) \\
Best TSFM $-$ reference & 0.006 (1/0/0) & 0.001 (0/1/0) & $-$0.058 (0/0/1) & $-$0.019 (0/0/1) \\
Median TSFM $-$ reference & 0.003 (0/1/0) & $-$0.010 (0/1/0) & $-$0.085 (0/0/1) & $-$0.033 (0/0/1) \\
\midrule
\multicolumn{5}{l}{\emph{AUC skill}} \\
LightGBM $-$ linear probe & 0.125 (6/0/0) & 0.039 (6/0/0) & 0.046 (6/0/0) & 0.041 (6/0/0) \\
LightGBM $-$ GBM probe & 0.081 (6/0/0) & 0.048 (6/0/0) & 0.054 (6/0/0) & 0.045 (6/0/0) \\
Linear probe $-$ raw logistic & 0.037 (6/0/0) & 0.013 (2/4/0) & 0.020 (4/2/0) & 0.019 (6/0/0) \\
Pretrained $-$ randomized & 0.019 (5/0/1) & 0.008 (2/4/0) & 0.048 (5/1/0) & 0.025 (5/1/0) \\
Best TSFM $-$ reference & 0.004 (0/1/0) & $-$0.005 (0/1/0) & $-$0.072 (0/0/1) & $-$0.025 (0/0/1) \\
Median TSFM $-$ reference & $-$0.003 (0/1/0) & $-$0.012 (0/1/0) & $-$0.097 (0/0/1) & $-$0.040 (0/0/1) \\
\bottomrule
\end{tabular}
}
\end{center}
\end{table}

\subsection{Probe training-size sensitivity}
\label{app:training-size}
We refit the linear probes of Chronos-2 and Timer on the seed-42 split with training caps from 10{,}000 to 150{,}000 windows, under both initializations.
We choose these two because Chronos-2 has a consistent pretrained advantage, whereas Timer is the backbone whose randomized features outperform its pretrained features on three datasets.

\begin{figure}[tb]
\begin{center}
\includegraphics[width=\textwidth]{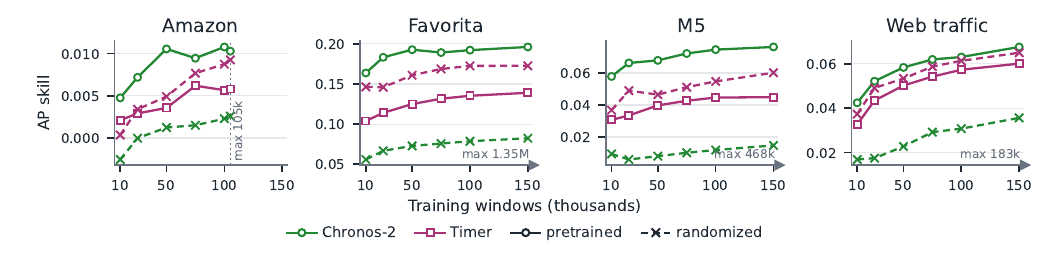}
\end{center}
\caption{AP skill of seed-42 linear probes on four datasets as a function of the number of training windows, under equal-window weighting.
Solid lines with hollow markers show pretrained backbones; dashed lines with crosses show randomized backbones.
Vertical dotted lines mark dataset maxima within the plotted range; right-pointing markers give maxima beyond 150{,}000 windows.
Amazon contains 105{,}189 eligible training windows, so the largest nominal cap does not add data beyond this point.
NYC taxi is omitted because its 22{,}278 available training windows cannot test the 100{,}000-window cap.
The points show fitted runs without intervals.}
\label{fig:probe-training-size}
\end{figure}

Raising the cap from 100{,}000 to 150{,}000 windows, or to all 105{,}189 Amazon training windows, changes AP skill by at most 0.0054 and leaves the sign of all eight pretrained-minus-randomized differences unchanged (Figure~\ref{fig:probe-training-size}).
This check covers two of the six backbones and does not establish that skill has saturated.

\subsection{GBM probe}
\label{app:readout-sensitivity}

Across the pretrained representations, the linear probe exceeds the GBM probe for all six backbones on Amazon, Favorita, M5, and NYC taxi, whereas GBM is higher for all six on web traffic.
All 60 paired intervals exclude zero in these directions under each weighting scheme; we do not identify the cause.
On web traffic, GBM narrows but does not close the gap to raw LightGBM (Section~\ref{sec:results-representation}).

\begin{figure}[tb]
\begin{center}
\includegraphics[width=\textwidth]{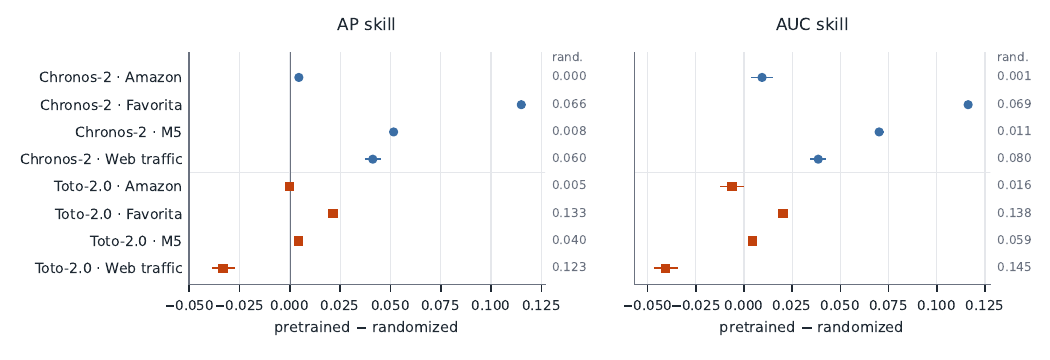}
\end{center}
\caption{Pretrained-minus-randomized difference in skill under the GBM probe for Chronos-2 (circles) and Toto-2.0 (squares), with pointwise 95\% paired bootstrap intervals over series; equal-window weighting. The right margin gives the randomized probe's absolute skill.}
\label{fig:randomized-gbm}
\end{figure}

GBM randomized controls cover Chronos-2 and Toto-2.0 on the four daily datasets; Chronos-2's eight differences are all above zero.
The pretrained-versus-randomized GBM comparison is probe-dependent for Toto-2.0 (Figure~\ref{fig:randomized-gbm}).
Favorita and M5 retain positive differences.
On Amazon, the AP interval spans zero and the AUC difference is $-0.0062$ $[-0.0124, -0.0002]$.
On web traffic, the randomized GBM probe attains 0.123 AP skill against 0.089 for the pretrained one ($-0.0332$ $[-0.0386, -0.0274]$), and both remain below raw LightGBM (0.177).
Under equal-series weighting, both Toto-2.0 differences on web traffic span zero (Appendix~\ref{app:weighting}).

\FloatBarrier
\section{Point-forecast diagnostics}
\label{app:diagnostics}

\subsection{Near-flat and near-zero point forecasts}
\label{app:degeneracy}

For each scored window, let $\hat x_{C+1},\ldots,\hat x_{C+64}$ denote its point forecast and $s$ the mean of its nonzero context values, where $s>0$ since every context contains at least two events (Section~\ref{sec:populations}).
We call the forecast near-flat if $\max_h\hat x_{C+h}-\min_h\hat x_{C+h}\leq 0.02s$ and near-zero if $\max_h|\hat x_{C+h}|\leq 0.01s$; every near-zero forecast is near-flat.
Table~\ref{tab:degeneracy} reports both shares, pooled over the test splits of each dataset.
Outside Amazon, the median forecasts are near-zero in 10.7--54.3\% and near-flat in 14.4--65.3\% of scored windows, whereas the forecasts of the five TSFMs that return a mean of samples or a model point output are near-zero in at most 1.7\% and near-flat in at most 3.9\%.
Exactly constant forecasts occur in at most 5.2\% of the scored windows of any TSFM--dataset pair, and exactly zero forecasts do not occur.
These shares describe the shape of the forecasts, not their ranking skill.

\begin{table}[tb]
\caption{Percentage of scored windows with near-flat point forecasts; percentages with near-zero forecasts are in parentheses. Median output: TSFMs whose point forecast is the predictive median (Table~\ref{tab:point-forecast}).}
\label{tab:degeneracy}
\begin{center}
\small
\begin{tabular}{lrrrrr}
\toprule
TSFM & Amazon & Favorita & M5 & NYC taxi & Web traffic \\
\midrule
\multicolumn{6}{l}{\textbf{Median output}} \\
Chronos-2 & 93.3 (84.4) & 24.0 (18.0) & 49.0 (37.8) & 29.6 (26.3) & 53.0 (36.3) \\
Chronos-Bolt & 90.4 (73.5) & 22.5 (18.2) & 37.5 (29.0) & 27.2 (20.4) & 38.9 (28.7) \\
Moirai-2.0 & 92.1 (81.6) & 28.8 (23.7) & 54.3 (45.2) & 56.0 (48.2) & 56.8 (43.0) \\
TimesFM-2.5 & 76.7 (66.0) & 14.4 (10.7) & 24.6 (17.9) & 25.9 (21.7) & 27.8 (21.2) \\
TimesFM-3 & 92.3 (85.9) & 25.7 (21.5) & 47.8 (40.5) & 36.9 (33.8) & 52.6 (43.2) \\
TiRex & 92.6 (89.5) & 25.8 (23.7) & 46.1 (42.5) & 31.2 (29.4) & 48.7 (42.1) \\
Toto-2.0 & 93.1 (91.4) & 33.5 (28.1) & 61.8 (53.9) & 39.5 (36.9) & 65.3 (54.3) \\
\midrule
\multicolumn{6}{l}{\textbf{Other point outputs}} \\
Moirai-1.1 & 0.0 (0.0) & 0.0 (0.0) & 0.0 (0.0) & 0.0 (0.0) & 0.1 (0.0) \\
Moirai-MoE & 0.0 (0.0) & 0.0 (0.0) & 0.0 (0.0) & 0.0 (0.0) & 0.0 (0.0) \\
Sundial & 1.4 (0.7) & 0.2 (0.1) & 0.1 (0.1) & 0.6 (0.3) & 0.4 (0.2) \\
TimeMoE & 32.4 (6.7) & 1.8 (0.8) & 1.2 (0.6) & 2.3 (1.7) & 2.0 (1.1) \\
Timer & 7.9 (1.5) & 0.7 (0.1) & 0.8 (0.1) & 3.9 (0.9) & 1.9 (0.4) \\
\bottomrule
\end{tabular}

\end{center}
\end{table}

\subsection{Rankings from quantiles}
\label{sec:quantiles}

The point forecast is one reduction of each TSFM's predictive distribution.
We rank positions by two other reductions of the quantile levels $\tau\in\{0.1,\dots,0.9\}$ and compare each with the point forecast on the same scored windows.

\paragraph{Approximate event probability.}
The first score is an approximate event probability $1-\hat F(0)$, computed for the ten TSFMs that output quantiles or samples (Table~\ref{tab:quantiles}).
The approximate score has lower skill than the point forecast in 48 of the 50 TSFM--dataset comparisons in AP skill and in 46 in AUC skill.
Its paired interval lies below zero in 47 AP and 42 AUC comparisons and includes zero in the others; none lies above zero.
Every comparison in which the score is not lower has a point-forecast skill below 0.006.
This conversion of quantiles therefore does not improve event ranking; the result concerns $1-\hat F(0)$, not the TSFMs' full predictive distributions.

\paragraph{Construction.}
The saved quantiles $q_{0.1}\le\dots\le q_{0.9}$ are monotone at every horizon position of every window.
We treat the forecast as a latent variable $\ry^*$ censored at zero, $\ry=\max(\ry^*,0)$, so that $P(\ry>0)=1-F^*(0)$ for the CDF $F^*$ of $\ry^*$.
We approximate $F^*$ by linear interpolation between the knots $(q_\tau,\tau)$.
If $q_{\tau_j}\le 0<q_{\tau_{j+1}}$ for consecutive levels $\tau_j<\tau_{j+1}$, this gives
\[
\hat F(0)=\tau_j+0.1\,\frac{-q_{\tau_j}}{q_{\tau_{j+1}}-q_{\tau_j}}.
\]
If $q_{0.1}>0$, we set $\hat F(0)=0$.
If $q_{0.9}\le 0$, we extend the segment between the knots at $q_{0.8}$ and $q_{0.9}$ linearly and cap $\hat F(0)$ at one.
The score $1-\hat F(0)$ is ranked within each window exactly like a point forecast.

\paragraph{Limits.}
The nine levels determine $\hat F(0)$ only when zero lies between $q_{0.1}$ and $q_{0.9}$; outside this range, $\hat F(0)$ follows from the extrapolation rule.
In particular, every position with $q_{0.1}>0$ receives the maximal score, so the score does not order these positions among themselves.
For Moirai-1.1, Moirai-MoE, and Sundial, the quantiles are empirical quantiles of 20 samples.
The score is therefore an approximation computed from the saved quantiles, not a native event probability of the TSFM.

\begin{table}[p]
\caption{Approximate event probability $1-\hat F(0)$ versus the point forecast for the ten TSFMs that output quantiles or samples, equal-window weighting. Point: skill of the point forecast (as in Tables~\ref{tab:complete-amazon}--\ref{tab:complete-web-traffic}); Approx.: skill of $1-\hat F(0)$; $\Delta$: Approx.\ minus Point, with 95\% paired bootstrap intervals over series. TimeMoE and Timer output neither.}
\label{tab:quantiles}
\begin{center}
\scriptsize
\setlength{\tabcolsep}{3pt}
\begin{tabular}{lrrr@{\ }lrrr@{\ }l}
\toprule
 & \multicolumn{4}{c}{AP skill} & \multicolumn{4}{c}{AUC skill} \\
\cmidrule(lr){2-5}\cmidrule(lr){6-9}
 & Point & Approx. & \multicolumn{2}{c}{$\Delta$ [95\% CI]} & Point & Approx. & \multicolumn{2}{c}{$\Delta$ [95\% CI]} \\
\midrule
\multicolumn{9}{l}{\textit{Amazon}} \\
Chronos-2 & 0.0057 & 0.0007 & $-$0.0051 & [$-$0.0068, $-$0.0034] & 0.0100 & $-$0.0012 & $-$0.0113 & [$-$0.0156, $-$0.0071] \\
Chronos-Bolt & 0.0055 & 0.0033 & $-$0.0022 & [$-$0.0035, $-$0.0010] & 0.0116 & 0.0047 & $-$0.0069 & [$-$0.0107, $-$0.0031] \\
Moirai-1.1 & 0.0004 & $-$0.0003 & $-$0.0007 & [$-$0.0022, 0.0009] & $-$0.0009 & $-$0.0006 & 0.0003 & [$-$0.0049, 0.0055] \\
Moirai-2.0 & 0.0050 & $-$0.0002 & $-$0.0052 & [$-$0.0067, $-$0.0038] & 0.0050 & $-$0.0047 & $-$0.0097 & [$-$0.0140, $-$0.0052] \\
Moirai-MoE & 0.0009 & 0.0014 & 0.0005 & [$-$0.0014, 0.0024] & $-$0.0007 & 0.0033 & 0.0040 & [$-$0.0014, 0.0093] \\
Sundial & 0.0030 & 0.0002 & $-$0.0029 & [$-$0.0046, $-$0.0012] & 0.0072 & 0.0042 & $-$0.0030 & [$-$0.0072, 0.0016] \\
TimesFM-2.5 & 0.0072 & 0.0023 & $-$0.0049 & [$-$0.0064, $-$0.0035] & 0.0103 & 0.0064 & $-$0.0039 & [$-$0.0070, $-$0.0007] \\
TimesFM-3 & 0.0057 & 0.0037 & $-$0.0020 & [$-$0.0034, $-$0.0006] & 0.0076 & 0.0068 & $-$0.0008 & [$-$0.0039, 0.0025] \\
TiRex & 0.0062 & 0.0022 & $-$0.0041 & [$-$0.0055, $-$0.0026] & 0.0056 & 0.0016 & $-$0.0040 & [$-$0.0074, $-$0.0007] \\
Toto-2.0 & 0.0052 & 0.0029 & $-$0.0023 & [$-$0.0035, $-$0.0010] & 0.0056 & 0.0073 & 0.0017 & [$-$0.0019, 0.0051] \\
\midrule
\multicolumn{9}{l}{\textit{Favorita}} \\
Chronos-2 & 0.1160 & 0.0292 & $-$0.0868 & [$-$0.0883, $-$0.0854] & 0.1106 & 0.0370 & $-$0.0736 & [$-$0.0749, $-$0.0723] \\
Chronos-Bolt & 0.0859 & 0.0553 & $-$0.0306 & [$-$0.0317, $-$0.0295] & 0.0816 & 0.0611 & $-$0.0205 & [$-$0.0215, $-$0.0194] \\
Moirai-1.1 & 0.0482 & 0.0145 & $-$0.0336 & [$-$0.0348, $-$0.0325] & 0.0478 & 0.0166 & $-$0.0312 & [$-$0.0323, $-$0.0302] \\
Moirai-2.0 & 0.1231 & 0.0434 & $-$0.0797 & [$-$0.0811, $-$0.0784] & 0.1163 & 0.0472 & $-$0.0691 & [$-$0.0704, $-$0.0680] \\
Moirai-MoE & 0.0661 & 0.0174 & $-$0.0487 & [$-$0.0500, $-$0.0474] & 0.0655 & 0.0271 & $-$0.0384 & [$-$0.0396, $-$0.0372] \\
Sundial & 0.1080 & 0.0028 & $-$0.1052 & [$-$0.1068, $-$0.1035] & 0.1052 & 0.0050 & $-$0.1002 & [$-$0.1016, $-$0.0987] \\
TimesFM-2.5 & 0.1254 & 0.0617 & $-$0.0637 & [$-$0.0649, $-$0.0625] & 0.1184 & 0.0733 & $-$0.0451 & [$-$0.0461, $-$0.0441] \\
TimesFM-3 & 0.1186 & 0.0476 & $-$0.0710 & [$-$0.0724, $-$0.0697] & 0.1029 & 0.0532 & $-$0.0497 & [$-$0.0509, $-$0.0485] \\
TiRex & 0.1107 & 0.0402 & $-$0.0705 & [$-$0.0719, $-$0.0691] & 0.0997 & 0.0446 & $-$0.0551 & [$-$0.0563, $-$0.0539] \\
Toto-2.0 & 0.0928 & 0.0341 & $-$0.0587 & [$-$0.0600, $-$0.0574] & 0.0821 & 0.0374 & $-$0.0447 & [$-$0.0459, $-$0.0434] \\
\midrule
\multicolumn{9}{l}{\textit{M5}} \\
Chronos-2 & 0.0453 & 0.0104 & $-$0.0349 & [$-$0.0365, $-$0.0333] & 0.0518 & 0.0165 & $-$0.0353 & [$-$0.0371, $-$0.0334] \\
Chronos-Bolt & 0.0329 & 0.0184 & $-$0.0145 & [$-$0.0158, $-$0.0131] & 0.0290 & 0.0201 & $-$0.0089 & [$-$0.0106, $-$0.0072] \\
Moirai-1.1 & 0.0173 & 0.0051 & $-$0.0121 & [$-$0.0134, $-$0.0110] & 0.0237 & 0.0069 & $-$0.0168 & [$-$0.0184, $-$0.0152] \\
Moirai-2.0 & 0.0561 & 0.0208 & $-$0.0353 & [$-$0.0369, $-$0.0339] & 0.0626 & 0.0257 & $-$0.0369 & [$-$0.0386, $-$0.0351] \\
Moirai-MoE & 0.0232 & 0.0063 & $-$0.0168 & [$-$0.0181, $-$0.0156] & 0.0301 & 0.0099 & $-$0.0202 & [$-$0.0219, $-$0.0185] \\
Sundial & 0.0365 & 0.0010 & $-$0.0354 & [$-$0.0371, $-$0.0338] & 0.0438 & 0.0026 & $-$0.0412 & [$-$0.0430, $-$0.0394] \\
TimesFM-2.5 & 0.0536 & 0.0247 & $-$0.0290 & [$-$0.0304, $-$0.0276] & 0.0574 & 0.0332 & $-$0.0243 & [$-$0.0256, $-$0.0230] \\
TimesFM-3 & 0.0458 & 0.0208 & $-$0.0249 & [$-$0.0264, $-$0.0234] & 0.0379 & 0.0226 & $-$0.0153 & [$-$0.0169, $-$0.0137] \\
TiRex & 0.0401 & 0.0130 & $-$0.0271 & [$-$0.0287, $-$0.0257] & 0.0410 & 0.0120 & $-$0.0289 & [$-$0.0306, $-$0.0274] \\
Toto-2.0 & 0.0277 & 0.0117 & $-$0.0160 & [$-$0.0174, $-$0.0146] & 0.0270 & 0.0141 & $-$0.0129 & [$-$0.0145, $-$0.0114] \\
\midrule
\multicolumn{9}{l}{\textit{NYC taxi}} \\
Chronos-2 & 0.2117 & 0.0449 & $-$0.1668 & [$-$0.1856, $-$0.1472] & 0.2572 & 0.0510 & $-$0.2062 & [$-$0.2268, $-$0.1839] \\
Chronos-Bolt & 0.2164 & 0.0988 & $-$0.1176 & [$-$0.1299, $-$0.1043] & 0.2710 & 0.1592 & $-$0.1119 & [$-$0.1287, $-$0.0946] \\
Moirai-1.1 & 0.0035 & 0.0012 & $-$0.0023 & [$-$0.0042, $-$0.0005] & 0.0051 & 0.0018 & $-$0.0033 & [$-$0.0075, 0.0009] \\
Moirai-2.0 & 0.1609 & 0.0714 & $-$0.0895 & [$-$0.1043, $-$0.0751] & 0.2173 & 0.1306 & $-$0.0867 & [$-$0.1012, $-$0.0722] \\
Moirai-MoE & 0.0388 & 0.0304 & $-$0.0084 & [$-$0.0124, $-$0.0048] & 0.0460 & 0.0413 & $-$0.0047 & [$-$0.0094, $-$0.0002] \\
Sundial & 0.1976 & 0.0265 & $-$0.1711 & [$-$0.1928, $-$0.1503] & 0.2906 & 0.0783 & $-$0.2123 & [$-$0.2317, $-$0.1932] \\
TimesFM-2.5 & 0.1897 & 0.0986 & $-$0.0912 & [$-$0.1019, $-$0.0801] & 0.2444 & 0.1623 & $-$0.0821 & [$-$0.0923, $-$0.0711] \\
TimesFM-3 & 0.1992 & 0.1113 & $-$0.0879 & [$-$0.0991, $-$0.0764] & 0.2583 & 0.1710 & $-$0.0874 & [$-$0.0994, $-$0.0748] \\
TiRex & 0.2059 & 0.1254 & $-$0.0805 & [$-$0.0889, $-$0.0717] & 0.2564 & 0.2025 & $-$0.0538 & [$-$0.0608, $-$0.0468] \\
Toto-2.0 & 0.1757 & 0.0769 & $-$0.0988 & [$-$0.1128, $-$0.0852] & 0.2217 & 0.1196 & $-$0.1022 & [$-$0.1159, $-$0.0879] \\
\midrule
\multicolumn{9}{l}{\textit{Web traffic}} \\
Chronos-2 & 0.0187 & 0.0074 & $-$0.0113 & [$-$0.0138, $-$0.0089] & 0.0175 & 0.0096 & $-$0.0079 & [$-$0.0108, $-$0.0050] \\
Chronos-Bolt & 0.0149 & 0.0093 & $-$0.0056 & [$-$0.0075, $-$0.0038] & 0.0113 & 0.0074 & $-$0.0038 & [$-$0.0062, $-$0.0015] \\
Moirai-1.1 & 0.0043 & 0.0018 & $-$0.0025 & [$-$0.0044, $-$0.0007] & 0.0046 & 0.0039 & $-$0.0007 & [$-$0.0032, 0.0019] \\
Moirai-2.0 & 0.0214 & 0.0091 & $-$0.0123 & [$-$0.0144, $-$0.0104] & 0.0192 & 0.0081 & $-$0.0111 & [$-$0.0140, $-$0.0082] \\
Moirai-MoE & 0.0037 & 0.0038 & 0.0001 & [$-$0.0020, 0.0022] & 0.0039 & 0.0048 & 0.0009 & [$-$0.0021, 0.0038] \\
Sundial & 0.0159 & 0.0006 & $-$0.0153 & [$-$0.0181, $-$0.0123] & 0.0120 & 0.0004 & $-$0.0117 & [$-$0.0146, $-$0.0086] \\
TimesFM-2.5 & 0.0218 & 0.0076 & $-$0.0142 & [$-$0.0163, $-$0.0124] & 0.0106 & 0.0022 & $-$0.0084 & [$-$0.0105, $-$0.0063] \\
TimesFM-3 & 0.0218 & 0.0104 & $-$0.0114 & [$-$0.0132, $-$0.0095] & 0.0173 & 0.0100 & $-$0.0073 & [$-$0.0095, $-$0.0049] \\
TiRex & 0.0201 & 0.0113 & $-$0.0088 & [$-$0.0107, $-$0.0070] & 0.0160 & 0.0114 & $-$0.0046 & [$-$0.0070, $-$0.0024] \\
Toto-2.0 & 0.0197 & 0.0102 & $-$0.0095 & [$-$0.0114, $-$0.0077] & 0.0139 & 0.0090 & $-$0.0049 & [$-$0.0075, $-$0.0024] \\
\bottomrule
\end{tabular}

\end{center}
\end{table}

\paragraph{Quantile average.}
The second score is the average of the quantiles, $\bar q=\frac{1}{9}\sum_{\tau}q_\tau$, at each position, which replaces the median as the point reduction.
We compute it for the seven TSFMs that forecast the median of quantiles they output natively.
For Moirai-1.1, Moirai-MoE, and Sundial, the quantiles are empirical quantiles of the same 20 samples whose mean is already the point forecast, so we omit them.
The average omits the tails beyond $q_{0.1}$ and $q_{0.9}$ and is therefore not the predictive mean.
Unlike $1-\hat F(0)$, it improves on the median forecast, with paired intervals above zero in 27 of 35 comparisons in AP skill and 23 in AUC skill (Table~\ref{tab:quantile-average}).
In AP skill, this holds for all seven TSFMs on Favorita, M5, and NYC taxi, but for only two on web traffic; one AP and two AUC intervals lie below zero, all on Amazon.
The largest gain of any TSFM over the reference rises from 0.031 to 0.057 AP skill, by Chronos-2 on NYC taxi.

For the three probed backbones that forecast the median (Chronos-2, Chronos-Bolt, and Toto-2.0), each probe exceeds the quantile average of its own backbone in 28 of 30 comparisons in point estimates; both exceptions are Chronos-2 on NYC taxi.
The strongest quantile average across TSFMs, however, exceeds most probes on M5 and NYC taxi under both metrics but exceeds the best probe only on NYC taxi in AP skill, while at least five of six probes remain ahead on Favorita and web traffic.
On Favorita, M5, and web traffic this strongest average comes from Moirai-2.0, which we did not probe.
Within-backbone gains of the probes therefore do not establish that probes exceed the strongest outputs across TSFMs.

\begin{table}[tb]
\caption{Average of the quantiles versus the point forecast, equal-window weighting.
Above: TSFMs, of the seven median forecasters, whose paired 95\% interval for the quantile average minus the point forecast lies above zero.
Median~$\Delta$: median of this difference over the seven TSFMs.
Max $-$ ref.: largest difference from the training-free reference, over the point forecasts of all 12 TSFMs (as in Table~\ref{tab:output-summary}) and over the quantile averages of the seven.
Probes: linear probes, of six, whose skill exceeds the largest quantile average, in point estimates.
Both maxima are descriptive test-set maxima.}
\label{tab:quantile-average}
\begin{center}
\scriptsize
\setlength{\tabcolsep}{3pt}
\begin{tabular}{lrrrrrrrrrr}
\toprule
 & \multicolumn{5}{c}{AP skill} & \multicolumn{5}{c}{AUC skill} \\
\cmidrule(lr){2-6}\cmidrule(lr){7-11}
 & & & \multicolumn{2}{c}{Max $-$ ref.} & & & & \multicolumn{2}{c}{Max $-$ ref.} & \\
\cmidrule(lr){4-5}\cmidrule(lr){9-10}
Dataset & Above & Median $\Delta$ & Point & Avg. & Probes & Above & Median $\Delta$ & Point & Avg. & Probes \\
\midrule
Amazon & 4/7 & 0.003 & 0.002 & 0.003 & 2/6 & 4/7 & 0.006 & $-$0.006 & $-$0.006 & 5/6 \\
Favorita & 7/7 & 0.012 & 0.023 & 0.041 & 5/6 & 6/7 & 0.012 & 0.005 & 0.028 & 5/6 \\
M5 & 7/7 & 0.011 & 0.013 & 0.032 & 1/6 & 5/7 & 0.011 & 0.001 & 0.031 & 2/6 \\
NYC taxi & 7/7 & 0.022 & 0.031 & 0.057 & 0/6 & 7/7 & 0.059 & 0.020 & 0.039 & 2/6 \\
Web traffic & 2/7 & 0.000 & 0.006 & 0.007 & 6/6 & 1/7 & 0.001 & 0.004 & 0.004 & 6/6 \\
\bottomrule
\end{tabular}

\end{center}
\end{table}

\subsection{Supervised readout of the outputs}
\label{app:output-readout}
We train the logistic learner of Appendix~\ref{app:impl-training} on the outputs of each probed backbone: its point forecast (six backbones) or its point forecast and nine quantiles (four backbones), divided by the context mean, with or without a $\log(1+\cdot)$ transform.
The readout is cross-fitted on the test windows with five folds grouped by series.
Within each training fold, 20\% of the series are held out to select the L2 penalty by AP skill.
On Favorita, M5, and web traffic, the probe exceeds every readout of its own backbone, with paired intervals above zero for all variants under both metrics (Table~\ref{tab:output-readout}).
On Amazon and NYC taxi, the median AP-skill gaps are at most 0.009.
The readout itself improves on the point forecast most on Favorita, by 0.038--0.056 AP skill.

The readout's training population, sample size, and penalty selection differ from the probe's, so these gaps are not matched comparisons.
Cross-fitting the raw-context logistic regression in the same way changes its AP skill by $-0.014$ to $-0.004$ on the daily datasets, comparable to the smallest M5 gap, and by $+0.086$ on NYC taxi (last row of each panel).
On NYC taxi, the cross-fitted learner is trained on windows of the 2024 test year, whereas the original one is trained on 2022.
The selected penalty is the smallest grid value in 30\% of readout fits (62\% on Favorita), and web-traffic labels include missing values (Appendix~\ref{app:web-missing}).

\begin{table}[tb]
\caption{Cross-fitted logistic readout of the outputs, equal-window weighting.
Probe $-$ readout: pretrained linear probe minus the readout of its own backbone's point forecast (six backbones) or point forecast and nine quantiles (four backbones).
Scaled: features divided by the context mean; log: $\log(1+\cdot)$ of the scaled features.
Readout $-$ point forecast: readout of the point forecast minus the forecast itself.
Raw logistic: cross-fitted minus original raw-context logistic regression, a check of the cross-fitting protocol.
Each cell gives the median over backbones and, in parentheses, how many paired 95\% intervals lie above zero, span zero, or lie below zero.}
\label{tab:output-readout}
\begin{center}
\scriptsize
\resizebox{\linewidth}{!}{\begin{tabular}{lrrrrr}
\toprule
Contrast & Amazon & Favorita & M5 & NYC taxi & Web traffic \\
\midrule
\multicolumn{6}{l}{\emph{AP skill}} \\
Probe $-$ readout, point, scaled & 0.006 (6/0/0) & 0.043 (6/0/0) & 0.027 (6/0/0) & 0.009 (4/2/0) & 0.040 (6/0/0) \\
Probe $-$ readout, point, log & 0.006 (6/0/0) & 0.021 (6/0/0) & 0.017 (6/0/0) & 0.005 (3/3/0) & 0.037 (6/0/0) \\
Probe $-$ readout, point + quantiles, scaled & 0.008 (4/0/0) & 0.035 (4/0/0) & 0.025 (4/0/0) & 0.005 (3/0/1) & 0.034 (4/0/0) \\
Probe $-$ readout, point + quantiles, log & 0.007 (4/0/0) & 0.018 (4/0/0) & 0.017 (4/0/0) & 0.003 (2/1/1) & 0.029 (4/0/0) \\
Readout $-$ point forecast, scaled & $-$0.001 (0/4/2) & 0.038 (6/0/0) & 0.004 (4/1/1) & 0.009 (4/2/0) & 0.008 (6/0/0) \\
Readout $-$ point forecast, log & $-$0.001 (0/3/3) & 0.056 (6/0/0) & 0.015 (6/0/0) & 0.010 (6/0/0) & 0.013 (6/0/0) \\
Raw logistic, cross-fitted $-$ stored & $-$0.004 (0/0/1) & $-$0.005 (0/0/1) & $-$0.014 (0/0/1) & 0.086 (1/0/0) & $-$0.005 (0/0/1) \\
\midrule
\multicolumn{6}{l}{\emph{AUC skill}} \\
Probe $-$ readout, point, scaled & 0.010 (5/1/0) & 0.052 (6/0/0) & 0.042 (6/0/0) & 0.029 (4/1/1) & 0.044 (6/0/0) \\
Probe $-$ readout, point, log & 0.010 (5/1/0) & 0.033 (6/0/0) & 0.030 (6/0/0) & 0.024 (4/1/1) & 0.040 (6/0/0) \\
Probe $-$ readout, point + quantiles, scaled & 0.015 (4/0/0) & 0.048 (4/0/0) & 0.036 (4/0/0) & 0.009 (3/1/0) & 0.040 (4/0/0) \\
Probe $-$ readout, point + quantiles, log & 0.013 (4/0/0) & 0.021 (4/0/0) & 0.024 (4/0/0) & 0.005 (3/0/1) & 0.034 (4/0/0) \\
Readout $-$ point forecast, scaled & 0.005 (3/3/0) & 0.031 (6/0/0) & 0.008 (5/0/1) & 0.036 (6/0/0) & 0.020 (6/0/0) \\
Readout $-$ point forecast, log & 0.005 (1/5/0) & 0.053 (6/0/0) & 0.021 (6/0/0) & 0.045 (6/0/0) & 0.024 (6/0/0) \\
Raw logistic, cross-fitted $-$ stored & $-$0.005 (0/1/0) & $-$0.005 (0/0/1) & $-$0.016 (0/0/1) & 0.151 (1/0/0) & $-$0.008 (0/0/1) \\
\bottomrule
\end{tabular}
}
\end{center}
\end{table}

\FloatBarrier

\FloatBarrier
\section{Representation diagnostics}
\label{app:timer}

We compare pretrained and randomized features of all six probed backbones on three diagnostics.
The effective rank \citep{roy2007effective} is $\exp\big(-\sum_i p_i \log p_i\big)$, where $p_i=\sigma_i/\sum_j \sigma_j$ are the normalized singular values of the centered feature matrix $\mH\in\R^{N\times D}$ of $N$ window representations; we report the ratio of randomized to pretrained effective rank.
The rate and pattern readouts are ridge regressions from the features to the event rate and the binary event pattern of the last 96 context steps.
They are fitted, tuned, and scored on disjoint 60/20/20\% partitions of the test series, and scored by held-out $R^2$, averaged over the 96 positions for the pattern.
Table~\ref{tab:geometry} reports the pretrained-minus-randomized differences.

\paragraph{Timer's reversal.}
On Favorita, the pretrained-minus-randomized difference is $-0.038$ $[-0.039, -0.036]$ AP skill.
On Amazon, the AUC interval lies below zero and the AP interval spans it.

\begin{table}[tb]
\caption{Representation diagnostics. Effective-rank ratio: randomized over pretrained effective rank, range over datasets. Rate and pattern: pretrained-minus-randomized held-out $R^2$ of linear readouts of the recent context event rate (range over datasets) and binary event pattern (per dataset).}
\label{tab:geometry}
\begin{center}
\small
\resizebox{\linewidth}{!}{%
\begin{tabular}{lrrrrrrr}
\toprule
 & & & \multicolumn{5}{c}{Pattern $R^2$ difference} \\
\cmidrule(lr){4-8}
Backbone & Eff.-rank ratio & Rate $R^2$ diff. & Amazon & Favorita & M5 & NYC taxi & Web traffic \\
\midrule
Chronos-2    & 0.41--0.72 & 0.188--0.603 & 0.145 & 0.336 & 0.331 & 0.389 & 0.370 \\
Chronos-Bolt & 2.74--4.02 & 0.489--0.739 & 0.130 & 0.352 & 0.281 & 0.272 & 0.328 \\
Sundial      & 0.60--0.70 & 0.301--0.564 & 0.080 & 0.186 & 0.142 & 0.126 & 0.253 \\
TimeMoE      & 0.15--0.20 & 0.020--0.245 & 0.132 & 0.203 & 0.234 & 0.307 & 0.186 \\
Timer        & 4.29--5.23 & 0.171--0.495 & $-$0.216 & $-$0.018 & $-$0.158 & $-$0.225 & $-$0.054 \\
Toto-2.0     & 0.74--0.93 & 0.171--0.504 & $-$0.054 & 0.059 & $-$0.034 & $-$0.070 & 0.129 \\
\bottomrule
\end{tabular}%
}
\end{center}
\end{table}

Pretraining reduces the effective rank of Timer's features more than that of any other backbone.
Chronos-Bolt shows a comparable reduction, yet its pretrained probe leads on all five datasets, so reduced rank alone does not produce the reversal.
Across the 30 backbone--dataset pairs, the effective-rank ratio correlates with the probe difference at Spearman $\rho=-0.44$, but in none of the five datasets separately at $p<0.05$ (six backbones each).
The pattern-readout difference tracks the probe difference more closely: $\rho=0.69$ pooled, $\rho=0.59$ without Timer, and $\rho=0.77$--$1.00$ within each dataset over the six backbones.
The rate-readout difference does not ($\rho=-0.04$).
Within Timer, the relation reverses ($\rho=-0.90$ over five datasets): NYC taxi has the largest pattern loss and the only pretrained lead.
These correlations are descriptive and uncorrected, and the within-dataset and within-Timer estimates rest on five or six points.

\end{document}